\documentclass{article}
\usepackage[preprint]{colm2026_conference}

\usepackage{microtype}
\ifdefined\XeTeXrevision\usepackage{fontspec}\fi
\usepackage{hyperref}
\usepackage{url}
\usepackage{booktabs}
\usepackage{array}
\usepackage{graphicx}
\usepackage{subcaption}
\usepackage{caption}
\usepackage{amsmath}
\usepackage{amssymb}
\usepackage{pifont}
\newcommand{\cmark}{\ding{51}}
\newcommand{\xmark}{\ding{55}}
\usepackage{xcolor}
\usepackage{enumitem}
\usepackage{placeins}
\usepackage{titletoc}
\usepackage[capitalize,noabbrev]{cleveref}

\usepackage{amsmath,amsfonts,bm}

\def\eqref#1{equation~\ref{#1}}

\def\1{\bm{1}}

\DeclareMathAlphabet{\mathsfit}{\encodingdefault}{\sfdefault}{m}{sl}
\SetMathAlphabet{\mathsfit}{bold}{\encodingdefault}{\sfdefault}{bx}{n}

\definecolor{darkblue}{rgb}{0, 0, 0.5}
\hypersetup{colorlinks=true, citecolor=darkblue, linkcolor=darkblue, urlcolor=darkblue}

\definecolor{wpaccent}{HTML}{B8573E}
\newcommand{\bigquestion}[1]{\begin{center}\vspace{-1pt}{\large\bfseries\itshape\color{wpaccent}#1}\vspace{-1pt}\end{center}}
\newcommand{\llmnote}[1]{\textit{#1}}
\newcommand{\wfact}[1]{\texttt{#1}}

\title{Weight-Space Programming Doesn't Need Transformers}

\author{Hyunjong Ok \& Jaeho Lee \\
Pohang University of Science and Technology (POSTECH) \\
\texttt{\{hyunjong.ok, jaeho.lee\}@postech.ac.kr} \\
\href{https://github.com/effl-lab/Weight-Space-Programming}{\textcolor{wpaccent}{Code}} \quad \href{https://wsp-no-transformer.github.io}{\textcolor{wpaccent}{Project page}}}

\begin{document}

\maketitle
\lhead{Preprint.}

\begin{figure}[!h]
\vspace{-16pt}
\centering
\includegraphics[width=0.88\linewidth]{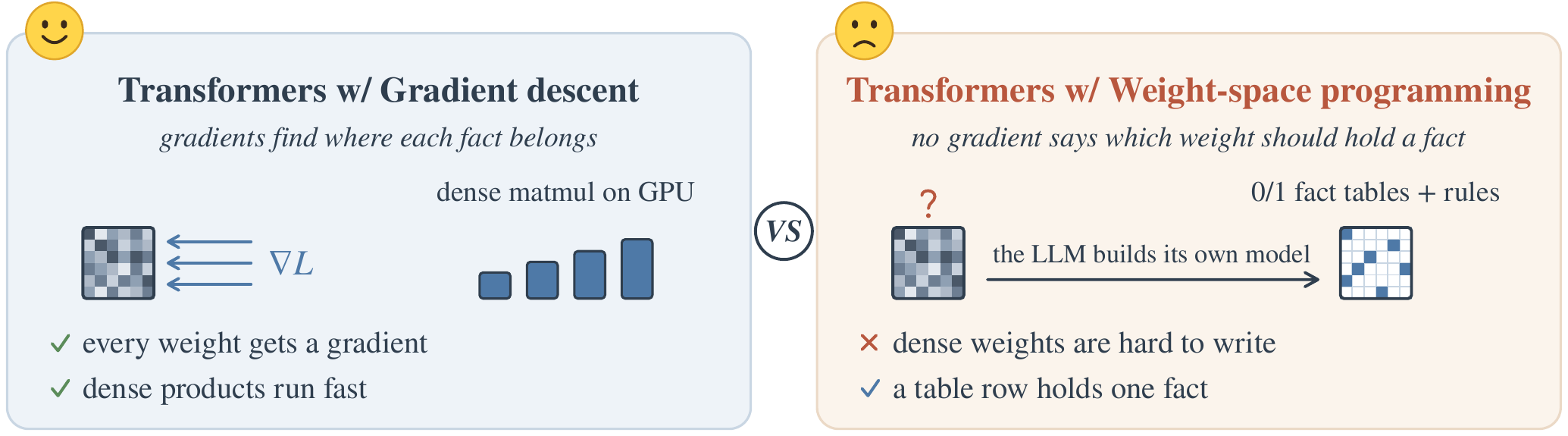}\\[3pt]
\includegraphics[width=0.98\linewidth]{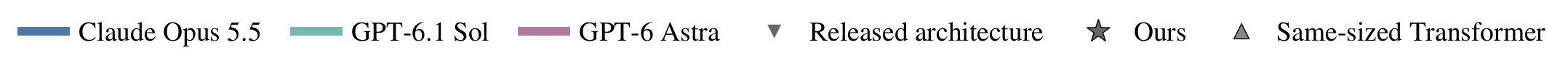}\\[2pt]
\captionsetup[subfigure]{skip=1pt}
\begin{subfigure}{0.508\linewidth}\centering
  \includegraphics[width=\linewidth]{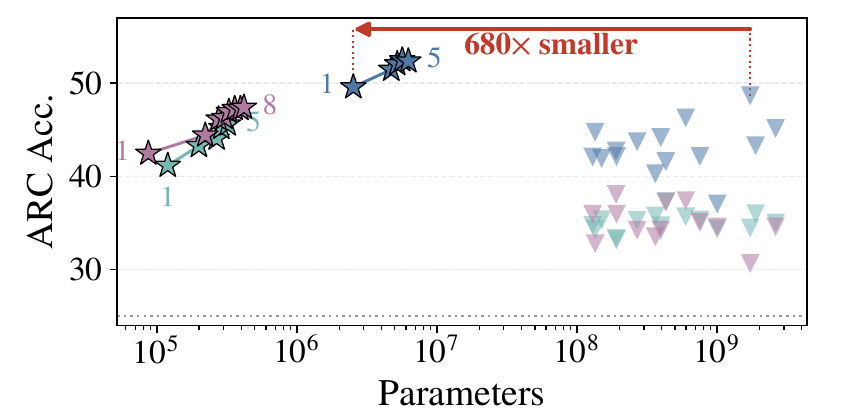}
  \subcaption{Size efficiency}\label{fig:teaser-size}
\end{subfigure}\hfill
\begin{subfigure}{0.481\linewidth}\centering
  \includegraphics[width=\linewidth]{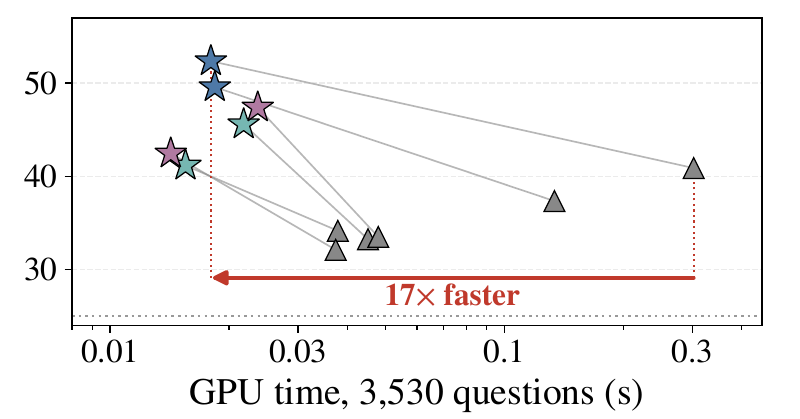}
  \subcaption{Speed}\label{fig:teaser-speed}
\end{subfigure}
\caption{\textbf{Transformers are inefficient for weight-space programming.} Gradient descent puts facts in dense weights; in weight-space programming, the LLM puts one fact per table row. Our associative-memory-based model is about 680$\times$ smaller than the most accurate released architecture that the same LLM wrote (a) and answers up to 17$\times$ faster than a same-sized Transformer (b). Numbers next to the stars give the generation.}\label{fig:teaser}
\end{figure}

\begin{abstract}
Standard neural architectures such as the Transformer were chosen because they train well with gradient descent and run fast on GPUs. In weight-space programming, a large language model (LLM) writes target weights by hand without gradient-based training, so those reasons no longer apply. Should we keep the Transformer? Given minimal guidance, ten labeled examples, and a sandbox, mainly on multiple-choice questions, we task three frontier LLMs with either generating the weights of a fixed architecture or designing the architecture themselves. When free to design, all three compile plain-text facts into a sparse table queried by a fixed procedure, reaching up to 54.4\% test accuracy. An LLM-designed model achieves higher test accuracy than the most accurate released Transformer that the same LLM wrote while using about 1,800$\times$ fewer parameters. Without gradients, they choose a table, in which the LLM can state what each weight means and can find and rewrite a wrong one. Each wrong answer points to one row, and removing a row changes about one other answer. From this, we propose an associative-memory-based model up to 17$\times$ faster on GPUs than a same-sized Transformer. We also test pretrained weights, other benchmarks, and tasks where models generate answers. Pretrained LLMs remain more accurate; the claim concerns which structure to build when writing weights.
\end{abstract}

\section{Introduction}

Transformers~\citep{vaswani2017attention} have been highly successful in part because their architecture is well suited to efficient gradient-based optimization on modern GPUs~\citep{hooker2020hardware, dao2024flashattention2}. However, this advantage does not necessarily carry over to weight-space programming (WSP), where a large language model (LLM) directly writes the weights of a separate target model without gradient-based training or data fitting~\citep{takehi2026weightspace}. This shift raises a central question.

\bigquestion{Should weight-space programming keep the Transformer?}

To answer this question, we provide minimal guidance, giving each LLM only ten labeled questions and an execution sandbox. Within this setup, we evaluate two regimes, (i) performing WSP on diverse released architectures (Transformers, state-space models~\citep{gu2022s4}, and RNNs~\citep{elman1990finding}) from random initialization, and (ii) allowing the LLM to choose and design the target architecture from scratch. We conduct our evaluation primarily on ARC multiple-choice science questions~\citep{clark2018arc} across three frontier LLMs, namely Claude Opus 5.5~\citep{anthropic2026opus55}, GPT-6.1 Sol~\citep{openai2026gpt61sol}, and GPT-6 Astra~\citep{openai2026gpt6astra}.

When free to choose the structure, the LLMs do not build Transformers; an LLM-designed model achieves higher ARC test accuracy than the most accurate released Transformer that the same LLM wrote while using about 1,800$\times$ fewer parameters.
We analyze in detail how these LLM-designed models are constructed, as building them differs from adjusting parameters via gradient descent. The LLM first expresses domain facts as plain-text sentences and writes a build script that converts each sentence into a row of a sparse table. A fixed lookup procedure then reads the table, matching tokens from the question and each candidate option against its rows to select the highest-scoring answer. All three LLMs converge on this design without using attention, MLPs, or dense embeddings.

This structure is well suited to writing weights without gradients because every prediction traces back to specific rows, and editing a row affects only the questions that rely on it. Empirically, in Claude Opus 5.5's models, retaining only the single best-matching row per option nearly matches full-model accuracy, and deleting a row alters roughly one prediction out of all test questions. By contrast, when Claude Opus 5.5 writes the same facts into a Transformer, it underperforms plain keyword search (BM25) over the raw sentences, which is consistent with interference between dense keys. Viewed as associative memories~\citep{zhong2025understanding,wang2026memory}, the sparse table pairs few-token keys with hard top-row selection, whereas the Transformer relies on dense, overlapping keys and soft attention.

However, an LLM-designed model scores every stored fact for every question, so inference slows down as the writer adds facts, and it launches many GPU kernels per question. Most of these models therefore run slower on a GPU than a same-sized Transformer. Building on our analysis, we propose an associative-memory-based model that retains the written table while reading only a fixed number of rows per question; it stays within about three percentage points of the LLM-designed models' accuracy while answering up to 17$\times$ faster on a GPU than the same-sized Transformer.

We test how well these findings generalize beyond random initialization and multiple-choice science questions, extending our evaluation to pretrained weights, math questions, open-ended generation, and vision tasks. Finally, while pretrained LLMs remain far more accurate overall, we clarify the practical boundaries of current WSP and outline key limitations and future directions for the paradigm.

\begin{table}[t]
\centering
\small
\setlength{\tabcolsep}{4pt}
\resizebox{\linewidth}{!}{%
\begin{tabular}{lcc}
\toprule
 & Gradient descent & Weight-space programming \\
\midrule
Weights set by & an optimizer, jointly & an LLM, line by line \\
What places a fact in the weights & the gradient & the writer \\
Hardware that builds the tensors & GPU & CPU \\
Needs a differentiable model & \cmark & \xmark \\
Needs an architecture built for gradient flow & \cmark & \xmark \\
Needs large train set & \cmark & \xmark \\
Weights traced to text a person can read & \xmark & \cmark \\
To change one fact & train again & edit one line \\
\midrule
Labeled ARC questions used & 3{,}358 & 10 \\
ARC accuracy in a two-layer Transformer (\%) & 29.8 & 36.1 \\
ARC accuracy in a written table (\%) & -- & 49.6--54.4 \\
\bottomrule
\end{tabular}}
\caption{\textbf{Gradient descent and weight-space programming ask different things of a model.} Upper rows compare how each way builds a model; bottom rows give Claude Opus 5.5 ARC accuracy, where the written-table range 49.6 to 54.4 covers its LLM-designed model and ours, generations 1 and 5.}
\label{tab:compare}
\end{table}

\section{Weight-space programming setup}\label{sec:setup}

We evaluate WSP on the four-option questions of the combined ARC-Easy and ARC-Challenge test sets~\citep{clark2018arc}, totaling 3,530 science questions. Three frontier LLMs (Claude Opus 5.5, GPT-6.1 Sol, and GPT-6 Astra), which we call \emph{writers}, operate under identical tools, sandbox environments, and task prompts. During a single run, each writer sees only ten ARC training examples, five each from ARC-Easy and ARC-Challenge, and sees no other ARC data and no test accuracy; in a lineage, it also receives the parent model and the parent's validation accuracy.

Under this setup, we compare two architectural regimes,
(i) \textbf{Fixed architectures}, where the writer writes weights over a random initialization of a released neural architecture, or writes every weight of a small two-layer baseline we provide with no starting weights (the \emph{given Transformer}); and
(ii) \textbf{LLM-designed models}, where the writer defines the structure itself via an architecture file, a build script, and plain-text knowledge files that are compiled into tensors on CPU.
Finally, we evaluate \emph{lineages}, where each generation refines the prior generation's top-performing validation model (\Cref{app:protocol} details the protocol, and \Cref{app:released} lists the released models).

\section{Written models as associative memories}\label{sec:structure}

\subsection{One kind of model from three LLMs}
\begin{figure}[t]
\centering
\captionsetup[subfigure]{justification=centering}
  \begin{subfigure}[b]{0.267\linewidth}
    \centering
    \includegraphics[width=\linewidth]{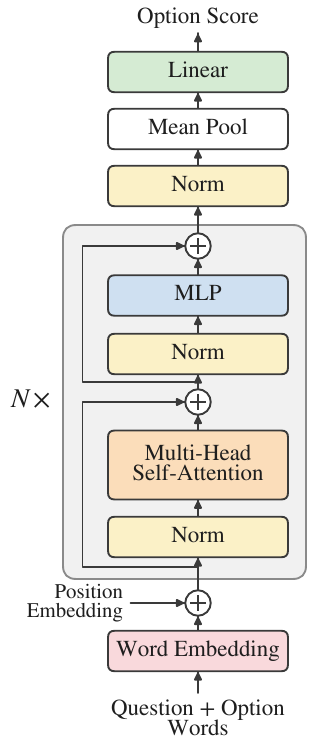}
    \subcaption{Given Transformer}\label{fig:arch-given}
  \end{subfigure}\hfill
  \begin{subfigure}[b]{0.323\linewidth}
    \centering
    \includegraphics[width=\linewidth]{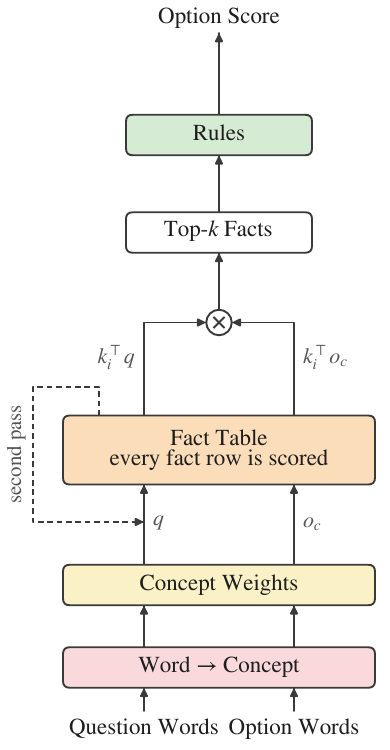}
    \subcaption{LLM-designed model}\label{fig:arch-designed}
  \end{subfigure}\hfill
  \begin{subfigure}[b]{0.36\linewidth}
    \centering
    \includegraphics[width=\linewidth]{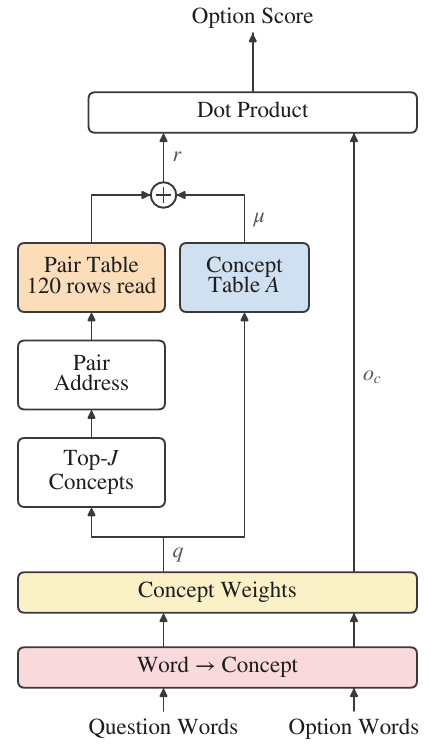}
    \subcaption{Associative-memory-based (ours)}\label{fig:arch-memory}
  \end{subfigure}
  \caption{\textbf{Stacked dense layers give way to written tables read once.} Colored blocks hold weights the LLM writes and white blocks are fixed operations; $q$ is the question's concept vector and $o_c$ holds the option concepts that the question lacks.}
  \label{fig:architecture}
\end{figure}

All three LLMs converge on the same kind of model, a sparse fact table queried by a fixed procedure (\cref{app:models} details all three writers). \Cref{fig:architecture} compares the given two-layer Transformer, Claude Opus 5.5's first-generation LLM-designed model,\footnote{Unless noted otherwise, single-model examples and statistics in the main text come from Claude Opus 5.5's first-generation models.} and our associative-memory-based model (\cref{sec:ours}). While every block in the Transformer's attention and MLP stack is a dense matrix, the LLM-designed model maps 361,150 words into 3,368 concepts (each grouping words of one meaning, with rarer concepts weighted higher) and matches the question and each option against a sparse fact table (4,288 rows, 29,502 non-zero entries). It keeps the top rows per option, runs a second pass expanded with concepts from the top-matching facts, and applies rules for negation, absolutes, and word order.

\subsection{What writing weights without gradients requires}\label{sec:building}
Without gradient descent (\cref{tab:compare}), each writer works in a loop, creating practice questions, running the model, inspecting failures, and adding a rule or fact row. This loop produced the mechanisms above, including higher weights on rare concepts, sign flips for negation, and the second pass for two-step recall. Debugging weights requires two properties.
\begin{itemize}[leftmargin=*, nosep]
    \item \textbf{Traceable answers.} Each prediction comes from a small, identifiable set of written lines, allowing the writer to locate the exact entry responsible for a mistake.
    \item \textbf{Local edits.} Adding, modifying, or removing a single written line changes only the predictions that depend on it, leaving all other predictions intact.
\end{itemize}
Claude Opus 5.5 states these requirements in its first-generation runs, noting \llmnote{Without training, the only way to get knowledge into the model is to write it, and short declarative facts are the cheapest form to write and check}, and \llmnote{A sparse retrieval model makes every number traceable to a sentence I wrote}. \Cref{sec:analysis} tests both properties.

\subsection{The LLM-designed model as an associative memory}
A Transformer is itself a stack of associative memories~\citep{zhong2025understanding,wang2026memory}, in which a query $x$ recalls values $v_i$ whose keys $k_i$ it resembles via
\begin{equation}
f(x) = \sum_{i=1}^{M} v_i\,\sigma\!\left(k_i^\top x\right).
\label{eq:am}
\end{equation}
Here $M$ is the number of memories and $\sigma$ is a selection function. It is the identity in a linear layer, an activation such as GELU in an MLP, whose first-matrix rows are the keys, and a softmax over context keys in self-attention. Because Transformer keys are dense and overlap, recall mixes in values from other memories, although a sharper $\sigma$ reduces interference.

The LLM-designed model instantiates the same memory with sparser choices (\cref{tab:memory_view}). Each memory $i$ is one written fact whose sparse concept vector $k_i$ serves as both key and value ($v_i = k_i$). With $q$ the question's concept vector and $o_c$ the concepts in option $c$ absent from the question, scoring option $c$ by matching $o_c$ against the best-supported fact gives
\begin{equation}
s_c = \max_{i}\;\left(k_i^\top q\right)\left(k_i^\top o_c\right),
\label{eq:designed}
\end{equation}
while the full model adds smaller weights from secondary rows (\cref{app:models}). Keys with roughly seven non-zero entries across 3,368 concepts rarely overlap, and hard top-$k$ selection discards whatever overlap remains.

\begin{table}[t]
\centering
\small
\setlength{\tabcolsep}{3pt}
\resizebox{\linewidth}{!}{%
\begin{tabular}{llll}
\toprule
 & Trained Transformer & LLM-designed & Associative-memory-based (ours) \\
\midrule
Memory unit & learned vector & written sentence & concept or concept pair \\
Key space & dense & sparse, 3{,}368 concepts & concept and pair indices \\
Selection $\sigma$ & GELU, softmax & top-$k$ rows & direct address \\
Set by & joint gradients & line by line & line by line \\
Cost per question & fixed by width & scales with facts & 120 pair rows and one product with $A$ \\
\bottomrule
\end{tabular}}
\caption{\textbf{All three models are associative memories differing in key space, selection, and construction.} Each column instantiates \cref{eq:am} for a two-layer Transformer trained by gradient descent, Claude Opus 5.5's first-generation model, and our model.}
\label{tab:memory_view}
\end{table}

\subsection{Our associative-memory-based model}\label{sec:ours}
LLM-designed models score every stored fact for every question (\cref{eq:designed}), so work per question grows with the number of written facts, and they launch many GPU kernels; on a GPU, most run slower than a same-sized Transformer (\cref{fig:speed}). We therefore design an associative-memory-based model that preserves line-by-line writing (\cref{fig:arch-memory}). Given our fixed forward pass along with its weight names and tensor shapes, the LLM writes a word list, plain-text files in line formats of its choice, and a build script that compiles them into two tables. Per question, the forward pass reads only 120 pair-table rows and computes a single product with concept table $A$, keeping per-question cost independent of fact count for a given concept vocabulary (\cref{sec:method} analyzes GPU speed and accuracy).

To populate the two tables, the LLM's build script compiles written lines such as cues with supported answers, terms with definition words, and parents with children. Concept table $A$ is a first-order memory holding how strongly concept $a$ alone recalls concept $b$, while the pair table is a second-order memory holding the concepts that a pair of concepts $(a, b)$, written with $a < b$, recalls together. Each pair maps directly to row $h(a, b) = (a p + b) \bmod H$ for a fixed prime $p$ and table height $H$, so the forward pass computes the row address and searches nothing. From a question's concept vector $q$, the $J = 16$ strongest concepts form the set $P(q)$ of 120 pairs. With $t_h$ denoting row $h$ of the pair table as a sparse vector over concepts, $\mu$ a written mixing weight, and $o_c$ the concepts in option $c$ absent from the question, recall and the option score are
\begin{equation}
r = \sum_{(a,b)\in P(q)} q_a\,q_b\;t_{h(a,b)} \;+\; \mu\,A^\top q,
\qquad
s_c = o_c^\top r,
\label{eq:assoc}
\end{equation}
instantiating \cref{eq:am} with linear keys in $A$ and pair keys in the pair table. While our forward pass fixes $p$ and $J$, the LLM sets $H$, the row width, and $\mu$. In Claude Opus 5.5's first-generation model, 40,312 words map to 3,805 concepts, table $A$ holds 2.06 million non-zero entries, and the pair table uses $H = 1{,}048{,}576$ rows, each storing up to 40 recalled concept IDs and weights. Although distinct pairs can share a row because concept pairs outnumber $H$, under 0.5\% of the pair table's 41.9 million weight slots are non-zero, and folding the table to 65,536 rows preserves accuracy (\cref{app:speed}). Under the parameter count of \cref{app:protocol}, this model has 2.53 million parameters (mostly in $A$), 2.7 times the 0.94 million of the first-generation LLM-designed model.

\subsection{An illustrative example}
\begin{figure}[t]
\centering
  \begin{subfigure}{\linewidth}
    \centering
    \includegraphics[width=\linewidth]{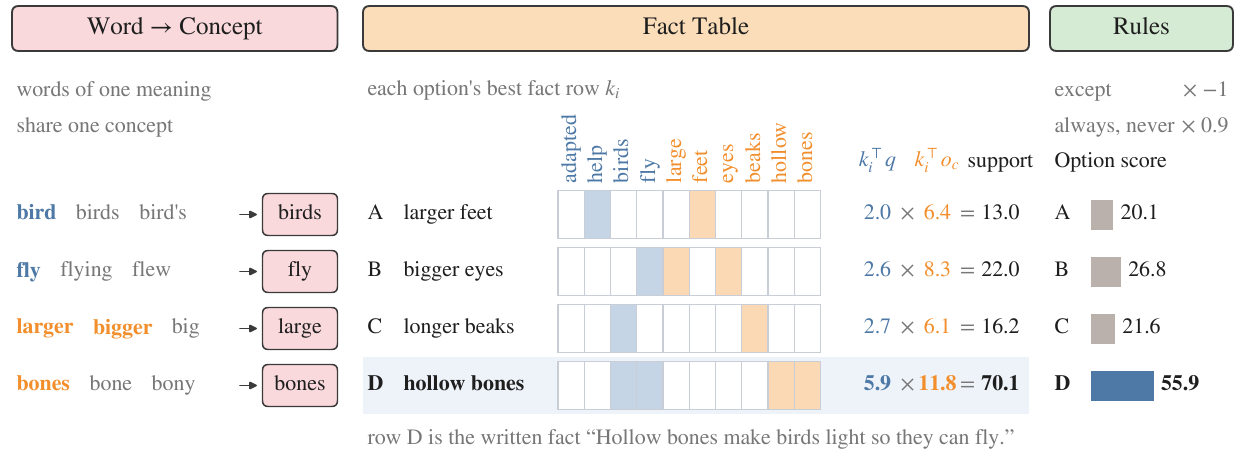}
    \subcaption{LLM-designed model}\label{fig:matrices-designed}
  \end{subfigure}\\[6pt]
  \begin{subfigure}{\linewidth}
    \centering
    \includegraphics[width=\linewidth]{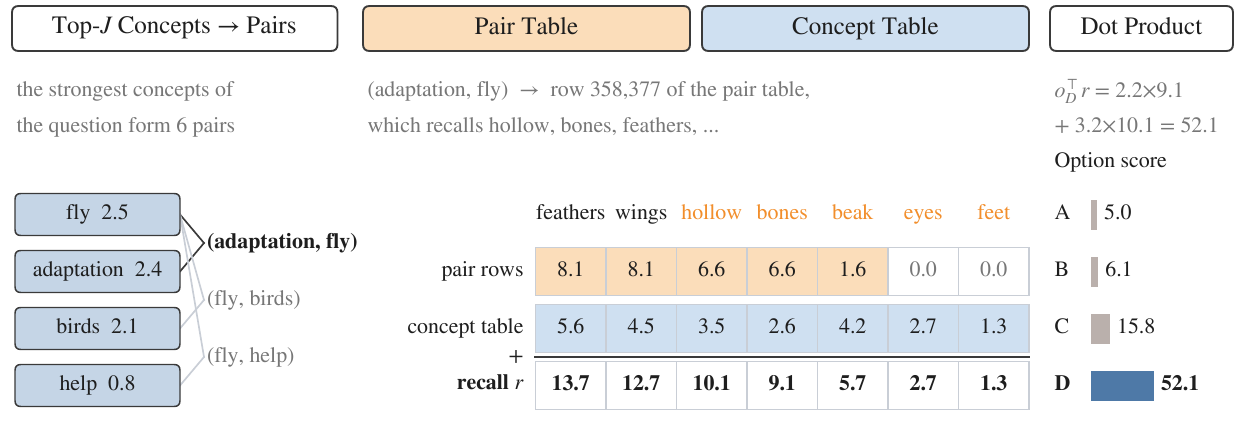}
    \subcaption{Associative-memory-based (ours)}\label{fig:matrices-memory}
  \end{subfigure}
  \caption{\textbf{In both models, the answer traces to written entries that a reader can name.} Panels follow an ARC test question through both Claude Opus 5.5 models, with blue question and orange option concepts. In (a), an option score scales the best fact's support and adds smaller shares of the next four facts, a window score, a second pass, and the rules, giving option D 55.9 from 70.1. In (b), the concept-table row includes $\mu = 1.5$, and entries are rounded to one decimal.}
  \label{fig:matrices}
\end{figure}

\Cref{fig:matrices} follows the unseen test question \emph{Which adaptation helps a bird fly?} In the LLM-designed model, related word forms such as \emph{bird}, \emph{birds}, and \emph{bird's}, or \emph{larger}, \emph{bigger}, and \emph{big}, share one concept, so the 4,288 written facts match whatever form the question uses. Each fact row is matched with both the question and an option, and their product gives the row's support. For \emph{hollow bones}, the written fact \wfact{Hollow bones make birds light so they can fly} reaches a support of 70 (with the next fact on hollow bones at 60), whereas the best rows for the three wrong options (\emph{larger feet}, \emph{bigger eyes}, and \emph{longer beaks}, matching facts about webbed feet, owl eyes, and a bird's beak) reach only 13 to 22. The answer therefore traces directly to written facts about hollow bones.

Our associative-memory-based model answers the same question without scanning the facts (\cref{fig:matrices-memory}). Among the six pairs formed by the question's four concepts, \emph{adaptation} and \emph{fly} address a pair-table row that recalls \emph{hollow} and \emph{bones}. Combined with concept table $A$, this gives \emph{hollow bones} the highest score (52.1 vs.\ 15.8 for the next option), and even without the pair table, concept table $A$ alone still ranks \emph{hollow bones} first (16.9 vs.\ 11.8).

\section{Results}\label{sec:arc}

Within weight-space programming, letting the LLM design the model structure outperforms fixing a standard architecture (\cref{tab:main_results}). Across all three LLMs, LLM-designed models achieve higher ARC test accuracy with far fewer parameters than released architectures, including Transformers, written from random initialization. Our associative-memory-based model (\cref{sec:ours}) retains this advantage over fixed architectures across all three LLMs. Refining the designed models across generations raises accuracy further, reaching 54.4\% in the strongest lineage.

\begin{table}[t]
\centering
\small
\setlength{\tabcolsep}{4pt}
\resizebox{\textwidth}{!}{%
\begin{tabular}{llrrrrr}
\toprule
 & Labeled & & LLM & \multicolumn{3}{c}{ARC test accuracy (\%)} \\
\cmidrule(lr){5-7}
Model & questions & Parameters & cost & Easy & Challenge & All \\
\midrule
\multicolumn{7}{l}{\textit{Gradient descent from random initialization}} \\
Best released architecture (Mamba-2 130M) & 3{,}358 & 129M & -- & 24.7 & 24.3 & 24.6 \\
Two-layer Transformer of an LLM-designed size & 3{,}358 & 3.33M & -- & 31.2 & 27.0 & 29.8 \\
Best small network of an LLM-designed size (MLP) & 3{,}358 & 3.34M & -- & 35.8 & 26.4 & 32.7 \\
\midrule
\multicolumn{7}{l}{\textit{Pretrained, fine-tuned}} \\
gemma-3-270m-it & 3{,}358 & 268M & -- & 43.3 & 30.9 & 39.2 \\
UltraBERT-150M & 3{,}358 & 190M & -- & 69.9 & 51.4 & 63.8 \\
Qwen3-0.6B & 3{,}358 & 596M & -- & 80.9 & 66.4 & 76.1 \\
Qwen3.5-0.8B & 3{,}358 & 752M & -- & 87.0 & 72.1 & 82.1 \\
\midrule
\multicolumn{7}{l}{\textit{Weight-space programming by Claude Opus 5.5}} \\
Released architectures (15), random initialization & 10 & 129M--2.6B & \$15 & 48.9 & 31.2 & 43.1 \\
Given two-layer Transformer (3 widths) & 10 & 0.08M--3.96M & \$16 & 40.0 & 28.2 & 36.1 \\
LLM-designed, generation 1 & 10 & 0.94M & \$29 & 59.0 & 34.2 & 50.8 \\
LLM-designed, generation 5 & 10 & 3.35M & \$144 & \textbf{62.6} & 37.8 & \textbf{54.4} \\
Associative-memory-based (ours), generation 1 & 10 & 2.53M & \$28 & 55.9 & 36.9 & 49.6 \\
Associative-memory-based (ours), generation 5 & 10 & 6.24M & \$96 & 59.0 & \textbf{38.9} & 52.4 \\
\midrule
\multicolumn{7}{l}{\textit{Weight-space programming by GPT-6.1 Sol}} \\
Released architectures (15), random initialization & 10 & 129M--2.6B & \$3 & 37.5 & 29.9 & 35.0 \\
Given two-layer Transformer & 10 & 0.03M--0.17M & \$2 & 41.3 & 26.2 & 36.3 \\
LLM-designed, generation 1 & 10 & 0.11M & \$6 & 51.0 & 31.6 & 44.6 \\
LLM-designed, generation 7 & 10 & 0.77M & \$45 & 54.1 & 33.7 & 47.4 \\
Associative-memory-based (ours), generation 1 & 10 & 0.12M & \$5 & 45.6 & 32.1 & 41.2 \\
Associative-memory-based (ours), generation 5 & 10 & 0.32M & \$36 & 50.8 & 35.1 & 45.6 \\
\midrule
\multicolumn{7}{l}{\textit{Weight-space programming by GPT-6 Astra}} \\
Released architectures (15), random initialization & 10 & 129M--2.6B & \$23 & 36.7 & 29.6 & 34.4 \\
Given two-layer Transformer & 10 & 0.31M & \$26 & 36.2 & 25.3 & 32.6 \\
LLM-designed, generation 1 & 10 & 0.16M & \$41 & 47.3 & 33.5 & 42.7 \\
LLM-designed, generation 5 & 10 & 0.55M & \$266 & 52.3 & 33.9 & 46.2 \\
Associative-memory-based (ours), generation 1 & 10 & 0.09M & \$41 & 47.3 & 32.7 & 42.5 \\
Associative-memory-based (ours), generation 8 & 10 & 0.42M & \$490 & 52.5 & 37.1 & 47.4 \\
\bottomrule
\end{tabular}}
\caption{\textbf{Written into a Transformer, the same LLM scores far lower than in a model it designs or in our associative-memory-based model.} LLM cost is one run for fixed architectures and the lineage to that generation otherwise. Multi-architecture rows average per-architecture means; bold marks the best value outside pretrained models.}
\label{tab:main_results}
\end{table}

Without pretraining, these written models also surpass gradient descent. Released architectures trained by gradient descent from the same random initializations stay near random guessing even on the full training set, and small networks trained from scratch at the scale of the LLM-designed models likewise fall well short. On ARC without pretraining, WSP thus reaches higher accuracy than training from scratch. Nonetheless, pretrained models fine-tuned on ARC remain far more accurate, with Qwen3.5-0.8B at 82.1\%, leaving a wide gap for WSP to narrow.

\section{Why the LLMs choose this structure}\label{sec:analysis}
Sparse tables satisfy both properties from \cref{sec:building} by tracing each answer to a single written line and limiting a line removal to far fewer changed answers than in a written Transformer. When compiled into a Transformer, however, the same content underperforms plain search over the written text, and edits no longer reliably stay local.

\subsection{Testing the two properties}\label{sec:trace}

\paragraph{Does an answer trace to one line?} To test whether each prediction relies on a single stored statement, we keep only the single best-matching fact for each option and drop every other score. In the LLM-designed model, the full model reaches 50.8\% test accuracy, and scoring each option from its single best-matching fact alone reaches 49.1\% (\cref{tab:controls}). Accuracy comes mostly from the best written fact for each option.

\begin{table}[t]
\centering
\small
\setlength{\tabcolsep}{4pt}
\begin{tabular}{lrr}
\toprule
Run & Non-zero values & ARC acc. \\
\midrule
LLM-designed, no restriction & 0.45M--0.57M & 50.7 \\
Practice questions forbidden & 0.05M--0.19M & 49.5 \\
Fact rows forbidden & 84M--392M & 45.6 \\
\midrule
Run 1 with no restriction, full model & 0.57M & 50.8 \\
\quad one best fact per option only & 0.57M & 49.1 \\
\bottomrule
\end{tabular}
\caption{\textbf{Forbidding practice questions costs about one point, forbidding fact rows costs about five points with over 100 times more values, and one best fact per option nearly reproduces the model.} Rows report ARC test accuracy (\%) for Claude Opus 5.5 first-generation LLM-designed models; upper rows average two runs.}
\label{tab:controls}
\end{table}

\paragraph{Does an edit stay local?} We remove one written line, rebuild the model, and compare it with the unchanged model on the test set for 40 random lines per model; rebuilding an unchanged model changes nothing. In the two LLM-designed models and the two associative-memory-based models, a removed line changes 0.5 to 1.8 answers on average (\cref{tab:line_ablation}). The 1.8 average comes from two word-to-concept vocabulary lines in one associative-memory-based model (23 and 30 answers), and without them that model changes 0.5. In the first written Transformer, a removed line changes 4.8 answers on average, several times more than in the table models. In the second written Transformer, its build script draws every vector code from one random stream, so one removed line changes almost every weight and 38\% of answers on average. The same test on the first-generation models of GPT-6.1 Sol and GPT-6 Astra (\cref{tab:line_ablation_writers}) gives 0.5 to 1.8 changed answers per removed line in their four LLM-designed models and 0.4 to 8.7 in their four associative-memory-based models. In the two runs above 1.8, at 5.7 and 8.7 changed answers, 53\% and 96\% of the changed answers come from lines that also alter the vocabulary. Their two written Transformers change 10.8 and 20.3 answers per removed line.

\begin{table}[t]
\centering
\small
\setlength{\tabcolsep}{4pt}
\resizebox{\linewidth}{!}{%
\begin{tabular}{lrrrrrrr}
\toprule
 & Non-zero & Weights & Answers & \multicolumn{2}{c}{Share a word with the line (\%)} & \multicolumn{2}{c}{Test accuracy (\%)} \\
\cmidrule(lr){5-6}\cmidrule(lr){7-8}
Model & values (M) & changed & changed & changed answers & all questions & before & after \\
\midrule
LLM-designed, run 1 & 0.57 & 166 & 0.9 & 75 & 16 & 50.7 & 50.7 \\
LLM-designed, run 2 & 0.45 & -- & 0.8 & 88 & 13 & 50.6 & 50.6 \\
Associative-memory-based (ours), run 1 & 2.53 & 6{,}203 & 1.8 & 85 & 10 & 49.3 & 49.3 \\
Associative-memory-based (ours), run 2 & 1.32 & 1{,}171 & 0.5 & 100 & 9 & 50.1 & 50.1 \\
Written Transformer, run 1 & 3.96 & 4{,}276 & 4.8 & 71 & 8 & 34.1 & 34.1 \\
Written Transformer, run 2 & 0.81 & nearly all & 1{,}350.7 & 7 & 7 & 34.7 & 35.3 \\
\bottomrule
\end{tabular}}
\caption{\textbf{A removed line changes 0.5 to 1.8 answers on average in a table model and more in a written Transformer.} Rows are the two first-generation runs of each model by Claude Opus 5.5. In the weights column, -- marks a model that renumbers its table columns when a line is removed, so its changed weights cannot be counted. The second written Transformer's build script draws every vector code from one random stream.}
\label{tab:line_ablation}
\end{table}

\subsection{Other explanations}\label{sec:controls}

\begin{table}[t]
\centering
\small
\setlength{\tabcolsep}{4pt}
\begin{tabular}{lrrr}
\toprule
Model & Written model & BM25 over its text & Difference \\
\midrule
Written two-layer Transformer & 36.1 & 44.1 & $-$8.0 \\
Written released architecture & 43.5 & 46.2 & $-$2.7 \\
\midrule
LLM-designed, generation 1 & 50.7 & 47.0 & $+$3.7 \\
LLM-designed, generation 5 & 54.1 & 48.4 & $+$5.7 \\
Associative-memory-based (ours), generation 1 & 49.9 & 48.4 & $+$1.5 \\
Associative-memory-based (ours), generation 5 & 52.6 & 49.1 & $+$3.5 \\
\bottomrule
\end{tabular}
\caption{\textbf{Written Transformers score below a plain search over their own text, and the table models score above it.} Columns report mean ARC test accuracy (\%) over the runs of each Claude Opus 5.5 model and for BM25 over the text written for each model.}
\label{tab:bm25}
\end{table}

\paragraph{Is it the content rather than the structure?} An alternative explanation is that the accuracy comes from the facts that the LLM writes, and the structure holding them does not matter. Given a Transformer, the LLM writes the same kind of fact list and compiles it into the weights. We run one plain search, BM25~\citep{robertson2009bm25}, over the text each LLM wrote for a model and compare it with that model's own accuracy. The six written two-layer Transformers of Claude Opus 5.5 all score below BM25 over their own text, by 8.0 points on average. The LLM-designed models and the associative-memory-based model score above it (\cref{tab:bm25}). Over all the text that GPT-6.1 Sol and GPT-6 Astra wrote, 8 of their 11 written two-layer Transformers score below BM25 and 13 of their 16 table models score above it (\cref{tab:bm25_writers}). In the terms of \cref{eq:am}, a written Transformer stores many memories under dense overlapping keys, and this result is consistent with interference. On the SciQ text generation task of \cref{sec:beyond}, where the model writes the answer in words, Claude Opus 5.5 measured the interference directly on its own practice questions. Its written Transformer holds about 2,800 answer words in 58 dimensions, and the same scoring rule gives an F1 of 0.37 without interference and 0.24 to 0.25 inside the Transformer.

\paragraph{Do the models copy test items or rely on made-up questions?} The text each LLM wrote overlaps the ARC test set about as often as it overlaps comparable questions that the LLM never saw, and deleting the overlapping lines barely changes test accuracy. Forbidding the LLM from writing practice questions of its own costs about one point (\cref{tab:controls}). \Cref{app:controls} gives the overlap analysis.

\paragraph{Is the table chosen because it is familiar?} One alternative reading of why the LLMs choose this structure remains open. An LLM knows retrieval as the classic way to answer questions without training, and one writer calls IDF retrieval \llmnote{the classic ARC baseline}, so our data do not separate a structure chosen because it suits writing from one chosen because it is known. We also ran a control without row tables. Forbidden to store facts as rows, Claude Opus 5.5 again built a memory read by key, using superposed random high-dimensional vectors with over 100 times more non-zero values (\cref{tab:controls}).

\section{Inference efficiency}\label{sec:method}

\subsection{Why LLM-designed models are slow on a GPU}\label{sec:slow}

\begin{figure}[t]
\centering
  \begin{subfigure}{0.49\linewidth}
    \centering
    \includegraphics[width=\linewidth]{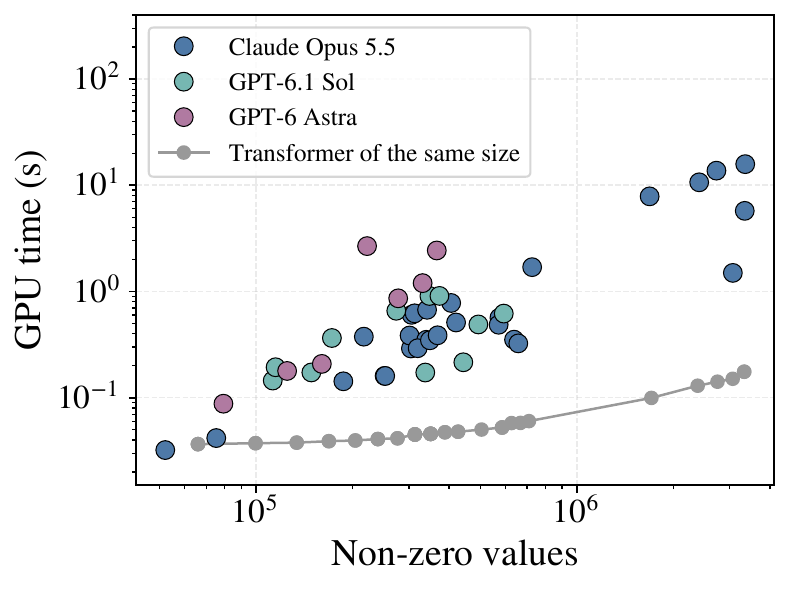}
    \subcaption{Fastest of eager, CUDA graph, and compiled GPU time vs.\ non-zero values}\label{fig:speed-a}
  \end{subfigure}\hfill
  \begin{subfigure}{0.49\linewidth}
    \centering
    \includegraphics[width=\linewidth]{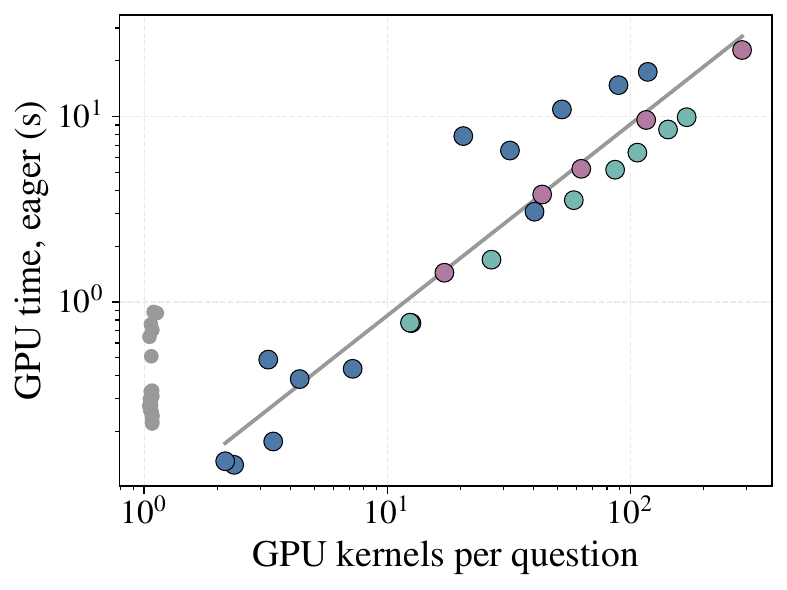}
    \subcaption{Eager GPU time vs.\ GPU kernels per question}\label{fig:speed-b}
  \end{subfigure}
  \caption{\textbf{GPU time follows kernel launches per question, and almost no LLM-designed model beats a Transformer of its size.} Each point is one model timed on the 3,530 ARC test questions on one RTX 3090. Colored circles are LLM-designed models, and gray circles are same-sized two-layer Transformers.}
  \label{fig:speed}
\end{figure}

A two-layer Transformer processes a batch of 64 questions in a few large matrix operations, averaging roughly one kernel launch per question, whereas LLM-designed models launch two to nearly three hundred (\cref{fig:speed}). In eager execution, log-log GPU time correlates more closely with kernels per question ($r = 0.95$) than with non-zero count. At each model's fastest mode, time tracks non-zero count more closely (correlation 0.84 against 0.64). Across 46 LLM-designed variants (\cref{fig:speed-a}), only one beats a same-sized Transformer at its fastest input, and the median is 9.4$\times$ slower.

These designs score every fact for every question, although in Claude Opus 5.5's lineage about one fact in seven shares a word with the question on average. They build an intermediate table with one entry per question, option, and fact, passing over that table several times. Over generations, the LLM adds scoring steps and rule modules, so kernels per question grow from 13 in Claude Opus 5.5's first generation to 118 in its fifth. Many designs build or transpose sparse tables inside the forward pass, which prevents capture as a CUDA graph, the optimization that removes kernel launch overhead. Cost grows with the number of facts times the number of scoring passes, and our associative-memory-based model (\cref{sec:ours}) removes this growth.

\subsection{What the architecture has to keep}

A faster design must keep what makes weights writable without gradients (a line as the unit, a nameable meaning for every entry, and sparse keys) and must stop scoring every fact for every question. \Cref{app:speed} maps each way WSP differs from gradient training to one choice of the associative-memory-based model.

Before designing an architecture ourselves, we asked Claude Opus 5.5 to rewrite its first-generation and fifth-generation LLM-designed models for a GPU, two runs each. The rewrites keep the starting accuracy and run faster than the models they start from. They move most sparse products to batched dense ones, but they still score every stored row for every question and remain 6 to 33 times slower than a same-sized Transformer at its fastest input. We therefore fix the forward pass ourselves and leave the weight writing to the LLM. The rewrites are detailed in \cref{app:speed}.

\subsection{Accuracy and speed}

Over five generations with Claude Opus 5.5, our ten models reach 49.6\% to 52.8\% ARC test accuracy, within about three percentage points of the LLM-designed models' 50.6\% to 54.4\% and well above a same-sized Transformer written by the same LLM (\cref{tab:speed_compare}). Per question, the forward pass reads 120 rows of the pair table and computes one product with the concept table $A$ regardless of line count, answering the 3,530 test questions on one RTX 3090 in 18 ms at the first- and fifth-generation sizes (\cref{fig:teaser-speed}). At its best input (sorted by length and cut to real length), a same-sized Transformer is 7 times slower at the first-generation size and 17 times slower at the fifth-generation size; \cref{app:speed} gives other input treatments. All ten share this forward pass. Nine replay as CUDA graphs in 18 to 20 ms, while one first-generation run stores the concept table $A$ as a sparse matrix that prevents graph replay and takes 70 ms.

\begin{table}[t]
\centering
\small
\setlength{\tabcolsep}{4pt}
\resizebox{\linewidth}{!}{%
\begin{tabular}{lrrrrrl}
\toprule
 & & & & \multicolumn{2}{c}{Same-sized Transformer} & \\
\cmidrule(lr){5-6}
 & Values & ARC acc. & GPU time (s) & ARC acc. & GPU time (s) & vs.\ Transformer \\
\midrule
LLM-designed, generation 1 & 0.57M & 50.8 & 0.566 & 34.5 & 0.052 & 11$\times$ slower \\
LLM-designed, generation 5 & 3.35M & 54.4 & 15.8 & 41.4 & 0.176 & 90$\times$ slower \\
\midrule
Associative-memory-based (ours), generation 1 & 2.53M & 49.6 & 0.018 & 37.4 & 0.134 & 7$\times$ faster \\
Associative-memory-based (ours), generation 5 & 6.24M & 52.4 & 0.018 & 40.9 & 0.302 & 17$\times$ faster \\
\bottomrule
\end{tabular}}
\caption{\textbf{Of these models, only the associative-memory-based model is faster than a Transformer of its size, and both kinds of written model are more accurate than that Transformer.} GPU time is for the 3,530 ARC test questions on one RTX 3090. The timed Transformer has the row's number of non-zero values and random weights, while the Transformer accuracy is the mean of two Transformers that Claude Opus 5.5 wrote at the same width, which hold more non-zero values because the writer chooses the vocabulary.}
\label{tab:speed_compare}
\end{table}

\subsection{Scaling and capacity}

In eager execution on synthetic tables with the shape of the written models, time per question stays at 0.012 ms on the GPU while the pair table grows from 3 million to 1 billion non-zero values, whereas a same-sized Transformer goes from 0.044 ms to 14.5 ms. Time instead grows with the number of concepts. With the concept table $A$ stored densely, as in the written models, the model becomes slower than a Transformer between about 15,000 and 32,000 concepts, while a variant that reads the concept table $A$ as row lists stays faster up to about 100,000 concepts. In the first-generation model, sharing rows of the pair table costs no accuracy down to 65,536 rows (49.6\% at 1,048,576 rows and 49.8\% at 65,536). \Cref{app:speed} gives the scaling curves and the capacity results.

\section{Further experiments}\label{sec:beyond}

This section asks how far our findings go beyond random initialization and multiple-choice science questions. Wherever the LLM designs the model, it keeps written tables and a fixed reader, adding rule modules when needed and adapting what a row holds to the task.

\subsection{WSP of pretrained weights}\label{sec:pretrained}

Starting from a released pretrained checkpoint instead of random initialization, WSP behaves differently. In most runs (48 of 87 across the three LLMs), the LLM changes under 1\% of the weights. Most of these edits calibrate how the answer letter is read out or move science-word embeddings toward written definitions, and accuracy stays within one point of the checkpoint's in 43 of these 48 runs. In 30 runs, the LLM overwrites the checkpoint, in 29 with a hand-written fact or rule model and one with a hand-built similarity circuit. Most overwrites (19 of 30) are on checkpoints near chance. The exceptions are GPT-6.1 Sol and GPT-6 Astra runs on the masked-language encoders UltraBERT and ModernBERT-large, which start above chance and lose accuracy when overwritten. The remaining nine keep the checkpoint and rewire it so that its own signal selects the answer letter. Checkpoints above 72\% are never overwritten. Per-checkpoint results are in \cref{app:further-pretrained}.

\subsection{Other benchmarks}\label{sec:transfer}

We run models written for ARC unchanged on other four-option benchmarks. On SciQ~\citep{welbl2017crowdsourcing}, science questions of the same level, the LLM-designed and associative-memory-based models both reach 47.7\% in generation 1 and 50.1\% in generation 5, within about four points of ARC. Accuracy is 30\% to 36\% on OpenBookQA~\citep{mihaylov2018openbookqa} and near 25\% chance on MMLU~\citep{hendrycks2021mmlu}. This pattern is consistent with Claude Opus 5.5 having written elementary science, matching SciQ rather than OpenBookQA or MMLU. \Cref{app:further-transfer} gives details.

\subsection{Text generation tasks}

We ask the LLM for models that produce text answers instead of choosing an option, extracting a passage span on SQuAD 1.1~\citep{rajpurkar2016squad} or writing an answer one word at a time on SciQ. The structure does not change in kind. For each task, Claude Opus 5.5 designs a new model of written lookup tables read by a fixed procedure, changing only what a row holds, with word lists and a reading procedure for SQuAD and answers with the words of questions about them for SciQ. Generation here means reading out a stored answer. In \cref{tab:open_tasks}, its LLM-designed model scores above the written Transformer and the same Transformer trained by gradient descent on all usable training items on both tasks. Details are in \cref{app:further-generation}.

\begin{table}[t]
\centering
\small
\setlength{\tabcolsep}{12pt}
\resizebox{\linewidth}{!}{%
\begin{tabular}{lcccc}
\toprule
 & LLM-designed & Written & Transformer trained & No-model \\
Task & model & Transformer & on all training items & reference \\
\midrule
SQuAD 1.1 (F1) & 42.3 & 33.2 & 16.1 & 18.8 \\
SciQ (F1) & 19.0 & 7.6 & 10.0 & 17.0 \\
\bottomrule
\end{tabular}}
\caption{\textbf{Without answer options, the LLM-designed models of Claude Opus 5.5 again score above the Transformers the same LLM wrote.} Written columns average two runs on 10,570 SQuAD 1.1 and 1,000 SciQ test items. The no-model reference returns the passage sentence or training-question answer with the highest word overlap.}
\label{tab:open_tasks}
\end{table}

\subsection{Several tasks at once}

We also test a single task statement that names several kinds of items at once, giving ten multiple-choice examples each from ARC, MMLU, and MathBench~\citep{liu2024mathbench} and ten writing items in total, five SQuAD passages and five CNN/DailyMail summaries~\citep{hermann2015teaching,see2017get}. All three LLMs build one model with a reader per kind on shared word tables, combining a fact table with exam rules for the questions, a span reader, and a copier of first sentences. On CNN/DailyMail, the first-sentence copiers of all three LLMs score below copying the first three sentences of the article (24.9 ROUGE-L). When no task is named and we ask only for a language model that continues text, runs differ in structure and score lower. Each writer guesses how the model will be tested and designs the structure for that guess. Full results are in \cref{app:further-tasks}.

\subsection{Images and text classification}\label{sec:images_text}
On CIFAR-10~\citep{krizhevsky2009learning} raw pixels, networks written by Claude Opus 5.5 reach only 21\% to 28\% across eight fixed architectures, against 10\% chance, which fits the view that a weight on a pixel has no meaning a writer can name, as detailed in \cref{app:images}. With meaning-level tokens from a pretrained DINOv2~\citep{oquab2024dinov2} encoder, a written table reaches 96.1\% without provided image labels. For each token, the writer sees five training images containing that token, names the object in each picture, and tallies classes per token. Gradient descent on the same tokens reaches about the same accuracy with 100 labels, so this accuracy comes from the encoder. On AG News text classification~\citep{zhang2015character}, the four architectures Claude Opus 5.5 wrote with no data reach about 84\%, which gradient descent from random initialization matches only with thousands of labeled articles, as detailed in \cref{app:text}.

\section{Limitations and future work}\label{sec:limitations_future}

\paragraph{Recall without learned representations.} The multiple-choice and SciQ table models, our associative-memory-based model, and the released architectures written from random initialization answer by recalling facts the LLM wrote. None chains facts across steps or composes a long answer. ARC-Challenge accuracy stays near 38\% while ARC-Easy reaches 63\% (\cref{tab:main_results}). Run unchanged, models written for ARC stay at chance on MMLU. On SQuAD the written models extract spans with word lists, and some written models compute, including arithmetic in the models written for several tasks at once or with no task named (\cref{sec:beyond}). Written models store what the LLM states and learn no representations of their own. A natural next step is weight-space programming that builds representations and acquires knowledge beyond the facts a writer can state.

\paragraph{Scope of the claim about Transformers.} Our finding that WSP does not need a Transformer applies to how current LLMs write weights from random initialization without running on data. In that setting, writing requires naming what each weight means, whereas Transformers are built to be trainable with gradient descent. If a later method writes weights beyond recall, the best architecture is an open question and should pass our controls (searching written text alone, counting weights actually written, and moving to other benchmarks). Three directions follow, namely writing calculation and comparison into weights, training by gradient descent from a written associative-memory-based model, and adding word order so that the model generates text.

\paragraph{Small-model speed boundary.} The associative-memory-based model removes the growth of inference time with the number of facts, but its time grows with the number of concepts. On a GPU, its speed advantage over a same-sized Transformer therefore holds up to the concept counts measured in \cref{sec:method} (between about 15,000 and 32,000 concepts with a dense concept table, and about 100,000 with row lists).

\paragraph{Hardware for lookup.} We measured inference on a single GPU model (an RTX 3090) built for dense matrix products, whereas per question our associative-memory-based model computes hashed addresses, reads a fixed number of table rows, and accumulates their sparse entries. In recommendation systems, near-memory processing and FPGA engines accelerate such table lookups~\citep{ke2020recnmp,jiang2021microrec}, and hash-based sparse CPU execution has outpaced GPUs on large workloads~\citep{chen2020slide}. Other designs replace inference multiplications with hashing and table lookups~\citep{blalock2021multiplying} or execute associative search directly in content-addressable memory~\citep{ni2019ferroelectric}. Whereas Transformers fit the dense-compute GPUs they co-evolved with~\citep{hooker2020hardware}, benchmarking written models on lookup-oriented hardware and co-designing the forward pass for it remain open directions.

\paragraph{Low-label text and warm-starting.} Although we have run no experiment in low-label domains such as medicine or personalization, our text results (\cref{app:text}) motivate using written networks when labels are scarce. Used to initialize gradient descent on 100 labeled AG News articles, a written network reaches up to 84.7\% accuracy, whereas training from random initialization on the same labels reaches at most 63.6\%.

\paragraph{Rubric-guided generational search.} Because an LLM can already design its own model structure and revise it across generations, the next step is an automated loop over generations that scores candidate models by a rubric combining accuracy per writing cost, model size, and inference speed.

\section{Related work}\label{sec:related}

Recent work shows that language models can directly read and write network weights without gradient training~\citep{takehi2026weightspace}. Earlier work constructs or compiles weights without a language model, studying transformer-like computation~\citep{weiss2021thinking}, compiling Transformers for interpretability~\citep{lindner2023tracr}, and hand-constructing fact-storing MLPs for Transformers~\citep{dugan2025constructing}. These works fix the architecture and ask what can be written into it, while we leave the structure to the LLM and compare it with the same LLM's Transformers.

Other work modifies an already trained model. Where low-rank adaptation updates a large language model by gradient training~\citep{hu2022lora}, model-editing methods locate and edit factual associations in GPT~\citep{meng2022rome} and extend those edits to mass-editing memory in a Transformer~\citep{meng2023memit}. Rather than retraining the model, these editing methods compute low-rank weight updates from forward-pass statistics and gradient-optimized target vectors on a trained model, whereas our LLMs start from nothing or from random initialization, apart from the pretrained edits of \cref{sec:pretrained}.

Prior work also searches over structures or programs and generates weights. With language models, \citet{chen2023evoprompting} perform code-level neural architecture search, \citet{nasir2024llmatic} add quality-diversity optimization, and \citet{romeraparedes2024funsearch} search over programs for mathematical discoveries, while \citet{ha2016hypernetworks} introduce hypernetworks as networks that generate weights. In the older line of fast weight programmers, a slow network trained by gradient descent outputs the weights of a second, fast network~\citep{schmidhuber1992fastweight}, and linear Transformers can be written as such fast weight programmers~\citep{schlag2021linear}. WSP shares only part of the name with fast weight programmers, since in WSP a language model writes the weights of a separate model itself and the written model receives no gradient training. These search and weight-generation methods train candidate structures with gradients or score many candidates by measured fitness on data, whereas our setup never trains candidate models with gradients. Each run sees only ten labeled example questions without benchmark accuracy, and across generations a lineage uses one validation accuracy per run only to pass the better of two runs to the next generation.

Although existing work reads Transformer layers as associative memories~\citep{zhong2025understanding,wang2026memory}, \citet{cheng2026conditional} argue that Transformers lack a native lookup primitive and simulate retrieval with computation. Whereas \citet{lample2019large} place product-key memory layers inside a Transformer as a learned key table read by top-k search, the Engram module of \citet{cheng2026conditional} adds constant-time lookup beside mixture-of-experts layers, hashing suffix N-grams of token ids to addresses in embedding tables pretrained by gradient descent and adding retrieved vectors, gated by the hidden state, to the residual stream. In our associative-memory-based model, the table read at a computed address is the whole model rather than a Transformer module, keys are concept pairs rather than token N-grams, and an LLM writes entries line by line without gradients.

\bibliography{custom}

@article{clark2018arc,
  title={Think you have Solved Question Answering? Try {ARC}, the {AI2} Reasoning Challenge},
  author={Clark, Peter and Cowhey, Isaac and Etzioni, Oren and Khot, Tushar and Sabharwal, Ashish and Schoenick, Carissa and Tafjord, Oyvind},
  journal={arXiv preprint arXiv:1803.05457},
  year={2018}
}

@inproceedings{vaswani2017attention,
  title={Attention Is All You Need},
  author={Vaswani, Ashish and Shazeer, Noam and Parmar, Niki and Uszkoreit, Jakob and Jones, Llion and Gomez, Aidan N. and Kaiser, Lukasz and Polosukhin, Illia},
  booktitle={Advances in Neural Information Processing Systems},
  volume={30},
  year={2017}
}

@inproceedings{lindner2023tracr,
  title={Tracr: Compiled Transformers as a Laboratory for Interpretability},
  author={Lindner, David and Kram{\'a}r, J{\'a}nos and Farquhar, Sebastian and Rahtz, Matthew and McGrath, Thomas and Mikulik, Vladimir},
  booktitle={Advances in Neural Information Processing Systems},
  volume={36},
  pages={37876--37899},
  year={2023}
}

@inproceedings{weiss2021thinking,
  title={Thinking Like Transformers},
  author={Weiss, Gail and Goldberg, Yoav and Yahav, Eran},
  booktitle={Proceedings of the 38th International Conference on Machine Learning},
  series={Proceedings of Machine Learning Research},
  volume={139},
  pages={11080--11090},
  year={2021}
}

@inproceedings{meng2022rome,
  title={Locating and Editing Factual Associations in {GPT}},
  author={Meng, Kevin and Bau, David and Andonian, Alex and Belinkov, Yonatan},
  booktitle={Advances in Neural Information Processing Systems},
  volume={35},
  pages={17359--17372},
  year={2022}
}

@inproceedings{meng2023memit,
  title={Mass-Editing Memory in a Transformer},
  author={Meng, Kevin and Sharma, Arnab Sen and Andonian, Alex and Belinkov, Yonatan and Bau, David},
  booktitle={International Conference on Learning Representations},
  year={2023}
}

@inproceedings{chen2023evoprompting,
  title={{EvoPrompting}: Language Models for Code-Level Neural Architecture Search},
  author={Chen, Angelica and Dohan, David M. and So, David R.},
  booktitle={Advances in Neural Information Processing Systems},
  volume={36},
  pages={7787--7817},
  year={2023}
}

@article{romeraparedes2024funsearch,
  title={Mathematical discoveries from program search with large language models},
  author={Romera-Paredes, Bernardino and Barekatain, Mohammadamin and Novikov, Alexander and Balog, Matej and Kumar, M. Pawan and Dupont, Emilien and Ruiz, Francisco J. R. and Ellenberg, Jordan S. and Wang, Pengming and Fawzi, Omar and Kohli, Pushmeet and Fawzi, Alhussein},
  journal={Nature},
  volume={625},
  number={7995},
  pages={468--475},
  year={2024},
  doi={10.1038/s41586-023-06924-6}
}

@inproceedings{nasir2024llmatic,
  title={{LLMatic}: Neural Architecture Search via Large Language Models and Quality Diversity Optimization},
  author={Nasir, Muhammad Umair and Earle, Sam and Togelius, Julian and James, Steven and Cleghorn, Christopher},
  booktitle={Proceedings of the Genetic and Evolutionary Computation Conference},
  pages={1110--1118},
  year={2024},
  publisher={ACM},
  doi={10.1145/3638529.3654017}
}

@article{oquab2024dinov2,
  title={{DINOv2}: Learning Robust Visual Features without Supervision},
  author={Oquab, Maxime and Darcet, Timoth{\'e}e and Moutakanni, Th{\'e}o and Vo, Huy and Szafraniec, Marc and others},
  journal={Transactions on Machine Learning Research},
  year={2024}
}

@techreport{krizhevsky2009learning,
  title={Learning Multiple Layers of Features from Tiny Images},
  author={Krizhevsky, Alex},
  institution={University of Toronto},
  year={2009},
  url={https://www.cs.toronto.edu/~kriz/learning-features-2009-TR.pdf}
}

@inproceedings{hu2022lora,
  title={{LoRA}: Low-Rank Adaptation of Large Language Models},
  author={Hu, Edward J. and Shen, Yelong and Wallis, Phillip and Allen-Zhu, Zeyuan and Li, Yuanzhi and Wang, Shean and Wang, Lu and Chen, Weizhu},
  booktitle={International Conference on Learning Representations},
  year={2022}
}

@article{allal2025smollm2,
  title={{SmolLM2}: When Smol Goes Big -- Data-Centric Training of a Small Language Model},
  author={Allal, Loubna Ben and Lozhkov, Anton and Bakouch, Elie and Bl{\'a}zquez, Gabriel Mart{\'\i}n and Penedo, Guilherme and others},
  journal={arXiv preprint arXiv:2502.02737},
  year={2025}
}

@article{yang2025qwen3,
  title={{Qwen3} Technical Report},
  author={Yang, An and Li, Anfeng and Yang, Baosong and Zhang, Beichen and Hui, Binyuan and others},
  journal={arXiv preprint arXiv:2505.09388},
  year={2025}
}

@article{gemmateam2025gemma3,
  title={Gemma 3 Technical Report},
  author={{Gemma Team} and Kamath, Aishwarya and Ferret, Johan and Pathak, Shreya and Vieillard, Nino and others},
  journal={arXiv preprint arXiv:2503.19786},
  year={2025}
}

@inproceedings{warner2025modernbert,
  title={Smarter, Better, Faster, Longer: A Modern Bidirectional Encoder for Fast, Memory Efficient, and Long Context Finetuning and Inference},
  author={Warner, Benjamin and Chaffin, Antoine and Clavi{\'e}, Benjamin and Weller, Orion and Hallstr{\"o}m, Oskar and Taghadouini, Said and Gallagher, Alexis and Biswas, Raja and Ladhak, Faisal and Aarsen, Tom and Adams, Griffin Thomas and Howard, Jeremy and Poli, Iacopo},
  booktitle={Proceedings of the 63rd Annual Meeting of the Association for Computational Linguistics (Volume 1: Long Papers)},
  pages={2526--2547},
  year={2025},
  publisher={Association for Computational Linguistics},
  doi={10.18653/v1/2025.acl-long.127}
}

@inproceedings{dao2024mamba2,
  title={Transformers are {SSMs}: Generalized Models and Efficient Algorithms Through Structured State Space Duality},
  author={Dao, Tri and Gu, Albert},
  booktitle={Proceedings of the 41st International Conference on Machine Learning},
  series={Proceedings of Machine Learning Research},
  volume={235},
  pages={10041--10071},
  year={2024}
}

@article{peng2025rwkv7,
  title={{RWKV-7} ``Goose'' with Expressive Dynamic State Evolution},
  author={Peng, Bo and Zhang, Ruichong and Goldstein, Daniel and Alcaide, Eric and Du, Xingjian and others},
  journal={arXiv preprint arXiv:2503.14456},
  year={2025}
}

@article{ha2016hypernetworks,
  title={{HyperNetworks}},
  author={Ha, David and Dai, Andrew and Le, Quoc V.},
  journal={arXiv preprint arXiv:1609.09106},
  year={2016}
}

@misc{takehi2026weightspace,
  title={Sparks of Weight-Space Programming: {LLMs} can natively read \& write neural network weights},
  author={Takehi, Rikiya and Teoh, Jayden and Zhang, Alex and Khattab, Omar},
  year={2026},
  month=nov,
  howpublished={\url{https://rikiyat.github.io/blog/weight-space-programming/}},
  note={Blog post, November 2, 2026}
}

@misc{qwenteam2026qwen35,
  title={{Qwen3.5}: Towards Native Multimodal Agents},
  author={{Qwen Team}},
  year={2026},
  month=feb,
  howpublished={\url{https://qwen.ai/blog?id=qwen3.5}},
  note={Model cards: \url{https://huggingface.co/Qwen/Qwen3.5-0.8B} and \url{https://huggingface.co/Qwen/Qwen3.5-2B}}
}

@article{park2026ultrabert,
  title={{UltraBERT}: Beyond the Limit of {BERT} Pre-Training},
  author={Park, Jungwoo and Lee, Taewhoo and Kang, Jaewoo},
  journal={arXiv preprint arXiv:2610.15028},
  year={2026}
}

@article{gemmateam2024gemma2,
  title={Gemma 2: Improving Open Language Models at a Practical Size},
  author={{Gemma Team} and Riviere, Morgane and Pathak, Shreya and Sessa, Pier Giuseppe and Hardin, Cassidy and others},
  journal={arXiv preprint arXiv:2408.00118},
  year={2024}
}

@article{dugan2025constructing,
  title={Constructing Efficient Fact-Storing {MLPs} for Transformers},
  author={Dugan, Owen and Garcia, Roberto and Junkins, Ronny and Liu, Jerry and Zinsley, Dylan and Eyuboglu, Sabri and Rudra, Atri and R{\'e}, Chris},
  journal={arXiv preprint arXiv:2512.00207},
  year={2025}
}

@article{schmidhuber1992fastweight,
  title={Learning to Control Fast-Weight Memories: An Alternative to Dynamic Recurrent Networks},
  author={Schmidhuber, J{\"u}rgen},
  journal={Neural Computation},
  volume={4},
  number={1},
  pages={131--139},
  year={1992},
  doi={10.1162/neco.1992.4.1.131}
}

@inproceedings{schlag2021linear,
  title={Linear Transformers Are Secretly Fast Weight Programmers},
  author={Schlag, Imanol and Irie, Kazuki and Schmidhuber, J{\"u}rgen},
  booktitle={Proceedings of the 38th International Conference on Machine Learning},
  series={Proceedings of Machine Learning Research},
  volume={139},
  pages={9355--9366},
  year={2021}
}

@article{zhong2025understanding,
  title={Understanding Transformer from the Perspective of Associative Memory},
  author={Zhong, Shu and Xu, Mingyu and Ao, Tenglong and Shi, Guang},
  journal={arXiv preprint arXiv:2505.19488},
  year={2025}
}

@misc{wang2026memory,
  title={You could have invented transformers from memory},
  author={Wang, Alex},
  year={2026},
  howpublished={\url{https://alexwang.ai/posts/associative-memory/}},
  note={Blog post, October 15, 2026}
}

@article{hooker2020hardware,
  title={The Hardware Lottery},
  author={Hooker, Sara},
  journal={Communications of the ACM},
  volume={64},
  number={12},
  pages={58--65},
  year={2021},
  doi={10.1145/3467017}
}

@inproceedings{dao2024flashattention2,
  title={{FlashAttention-2}: Faster Attention with Better Parallelism and Work Partitioning},
  author={Dao, Tri},
  booktitle={The Twelfth International Conference on Learning Representations},
  year={2024}
}

@misc{anthropic2026opus55,
  title={Introducing {Claude Opus 5.5}},
  author={{Anthropic}},
  year={2026},
  howpublished={\url{https://www.anthropic.com/news/claude-opus-5-5}},
  note={September 22, 2026}
}

@misc{openai2026gpt6astra,
  title={{GPT-6 Astra} System Card},
  author={{OpenAI}},
  year={2026},
  howpublished={\url{https://deploymentsafety.openai.com/gpt-6-astra}},
  note={September 3, 2026}
}

@misc{openai2026gpt61sol,
  title={Addendum to {GPT-6 Astra} System Card: {GPT-6.1 Sol}},
  author={{OpenAI}},
  year={2026},
  howpublished={\url{https://deploymentsafety.openai.com/gpt-6-1-sol}},
  note={September 29, 2026}
}

@article{elman1990finding,
  title={Finding Structure in Time},
  author={Elman, Jeffrey L.},
  journal={Cognitive Science},
  volume={14},
  number={2},
  pages={179--211},
  year={1990},
  doi={10.1207/s15516709cog1402_1}
}

@inproceedings{gu2022s4,
  title={Efficiently Modeling Long Sequences with Structured State Spaces},
  author={Gu, Albert and Goel, Karan and R{\'e}, Christopher},
  booktitle={International Conference on Learning Representations},
  year={2022}
}

@article{cheng2026conditional,
  title={Conditional Memory via Scalable Lookup: A New Axis of Sparsity for Large Language Models},
  author={Cheng, Xin and Tian, Rui and Zeng, Wangding and Dai, Damai and Chen, Qinyu and Wang, Bingxuan and Xie, Zhenda and Huang, Kezhao and Yu, Xingkai and Deng, Chengqi and Zhou, Shangyan and Zhao, Chenggang and Hao, Zhewen and Li, Yukun and Zhang, Han and Zhang, Zhengyan and Wei, Yixu and Xu, M. Y and Zhang, Huishuai and Zhao, Dongyan and Liang, Wenfeng},
  journal={arXiv preprint arXiv:2601.07372},
  year={2026}
}

@inproceedings{welbl2017crowdsourcing,
  title={Crowdsourcing Multiple Choice Science Questions},
  author={Welbl, Johannes and Liu, Nelson F. and Gardner, Matt},
  booktitle={Proceedings of the 3rd Workshop on Noisy User-generated Text},
  pages={94--106},
  year={2017}
}

@inproceedings{mihaylov2018openbookqa,
  title={Can a Suit of Armor Conduct Electricity? A New Dataset for Open Book Question Answering},
  author={Mihaylov, Todor and Clark, Peter and Khot, Tushar and Sabharwal, Ashish},
  booktitle={Proceedings of the 2018 Conference on Empirical Methods in Natural Language Processing},
  pages={2381--2391},
  year={2018}
}

@inproceedings{hendrycks2021mmlu,
  title={Measuring Massive Multitask Language Understanding},
  author={Hendrycks, Dan and Burns, Collin and Basart, Steven and Zou, Andy and Mazeika, Mantas and Song, Dawn and Steinhardt, Jacob},
  booktitle={International Conference on Learning Representations},
  year={2021}
}

@inproceedings{rajpurkar2016squad,
  title={{SQuAD}: 100,000+ Questions for Machine Comprehension of Text},
  author={Rajpurkar, Pranav and Zhang, Jian and Lopyrev, Konstantin and Liang, Percy},
  booktitle={Proceedings of the 2016 Conference on Empirical Methods in Natural Language Processing},
  pages={2383--2392},
  year={2016}
}

@inproceedings{hermann2015teaching,
  title={Teaching Machines to Read and Comprehend},
  author={Hermann, Karl Moritz and Kocisky, Tomas and Grefenstette, Edward and Espeholt, Lasse and Kay, Will and Suleyman, Mustafa and Blunsom, Phil},
  booktitle={Advances in Neural Information Processing Systems},
  volume={28},
  year={2015}
}

@inproceedings{see2017get,
  title={Get To The Point: Summarization with Pointer-Generator Networks},
  author={See, Abigail and Liu, Peter J. and Manning, Christopher D.},
  booktitle={Proceedings of the 55th Annual Meeting of the Association for Computational Linguistics (Volume 1: Long Papers)},
  pages={1073--1083},
  year={2017}
}

@inproceedings{zhang2015character,
  title={Character-level Convolutional Networks for Text Classification},
  author={Zhang, Xiang and Zhao, Junbo and LeCun, Yann},
  booktitle={Advances in Neural Information Processing Systems},
  volume={28},
  year={2015}
}

@inproceedings{liu2024mathbench,
  title={{MathBench}: Evaluating the Theory and Application Proficiency of {LLMs} with a Hierarchical Mathematics Benchmark},
  author={Liu, Hongwei and Zheng, Zilong and Qiao, Yuxuan and Duan, Haodong and Fei, Zhiwei and Zhou, Fengzhe and Zhang, Wenwei and Zhang, Songyang and Lin, Dahua and Chen, Kai},
  booktitle={Findings of the Association for Computational Linguistics: ACL 2024},
  pages={6884--6915},
  year={2024}
}

@article{robertson2009bm25,
  title={The Probabilistic Relevance Framework: {BM25} and Beyond},
  author={Robertson, Stephen and Zaragoza, Hugo},
  journal={Foundations and Trends in Information Retrieval},
  volume={4},
  number={1--2},
  pages={1--174},
  year={2009},
  doi={10.1561/1500000019}
}

@inproceedings{ke2020recnmp,
  title={{RecNMP}: Accelerating Personalized Recommendation with Near-Memory Processing},
  author={Ke, Liu and Gupta, Udit and Cho, Benjamin Youngjae and Brooks, David and Chandra, Vikas and Diril, Utku and others},
  booktitle={2020 ACM/IEEE 47th Annual International Symposium on Computer Architecture (ISCA)},
  pages={790--803},
  year={2020},
  doi={10.1109/ISCA45697.2020.00070}
}

@inproceedings{jiang2021microrec,
  title={{MicroRec}: Efficient Recommendation Inference by Hardware and Data Structure Solutions},
  author={Jiang, Wenqi and He, Zhenhao and Zhang, Shuai and Preu{\ss}er, Thomas B. and Zeng, Kai and Feng, Liang and Zhang, Jiansong and Liu, Tongxuan and Li, Yong and Zhou, Jingren and Zhang, Ce and Alonso, Gustavo},
  booktitle={Proceedings of Machine Learning and Systems},
  volume={3},
  pages={845--859},
  year={2021}
}

@inproceedings{chen2020slide,
  title={{SLIDE}: In Defense of Smart Algorithms over Hardware Acceleration for Large-Scale Deep Learning Systems},
  author={Chen, Beidi and Medini, Tharun and Farwell, James and Gobriel, Sameh and Tai, Charlie and Shrivastava, Anshumali},
  booktitle={Proceedings of Machine Learning and Systems},
  volume={2},
  pages={291--306},
  year={2020}
}

@inproceedings{blalock2021multiplying,
  title={Multiplying Matrices Without Multiplying},
  author={Blalock, Davis and Guttag, John},
  booktitle={Proceedings of the 38th International Conference on Machine Learning},
  series={Proceedings of Machine Learning Research},
  volume={139},
  pages={992--1004},
  year={2021}
}

@article{ni2019ferroelectric,
  title={Ferroelectric ternary content-addressable memory for one-shot learning},
  author={Ni, Kai and Yin, Xunzhao and Laguna, Ann Franchesca and Joshi, Siddharth and D{\"u}nkel, Stefan and Trentzsch, Martin and M{\"u}ller, Johannes and Beyer, Sven and Niemier, Michael and Hu, Xiaobo Sharon and Datta, Suman},
  journal={Nature Electronics},
  volume={2},
  number={11},
  pages={521--529},
  year={2019},
  doi={10.1038/s41928-019-0321-3}
}

@article{chowdhery2022palm,
  title={{PaLM}: Scaling Language Modeling with Pathways},
  author={Chowdhery, Aakanksha and Narang, Sharan and Devlin, Jacob and Bosma, Maarten and Mishra, Gaurav and Roberts, Adam and others},
  journal={arXiv preprint arXiv:2204.02311},
  year={2022}
}

@article{brown2020language,
  title={Language Models are Few-Shot Learners},
  author={Brown, Tom B. and Mann, Benjamin and Ryder, Nick and Subbiah, Melanie and Kaplan, Jared and Dhariwal, Prafulla and others},
  journal={arXiv preprint arXiv:2005.14165},
  year={2020}
}

@article{lample2019large,
  title={Large Memory Layers with Product Keys},
  author={Lample, Guillaume and Sablayrolles, Alexandre and Ranzato, Marc'Aurelio and Denoyer, Ludovic and Jégou, Hervé},
  journal={arXiv preprint arXiv:1907.05242},
  year={2019}
}

@article{gunasekar2023textbooks,
  title={Textbooks Are All You Need},
  author={Gunasekar, Suriya and Zhang, Yi and Aneja, Jyoti and Teodoro Mendes, Caio César and Del Giorno, Allie and Gopi, Sivakanth and others},
  journal={arXiv preprint arXiv:2306.11644},
  year={2023}
}

@article{singh2024evaluation,
  title={Evaluation data contamination in {LLMs}: how do we measure it and (when) does it matter?},
  author={Singh, Aaditya K. and Kocyigit, Muhammed Yusuf and Poulton, Andrew and Esiobu, David and Lomeli, Maria and Szilvasy, Gergely and others},
  journal={arXiv preprint arXiv:2411.03923},
  year={2024}
}
\bibliographystyle{colm2026_conference}

\clearpage
\appendix
\setcounter{topnumber}{4}\setcounter{totalnumber}{6}
\renewcommand{\topfraction}{0.9}\renewcommand{\textfraction}{0.07}\renewcommand{\floatpagefraction}{0.8}
\makeatletter\renewenvironment{table}[1][]{\@float{table}[!htbp]}{\end@float}\renewenvironment{figure}[1][]{\@float{figure}[!htbp]}{\end@float}\makeatother
\section*{Appendix contents}
\startcontents[appendix]
\printcontents[appendix]{l}{1}{\setcounter{tocdepth}{1}}
\clearpage
\section{WSP protocol}\label{app:protocol}

\begin{figure}[t]
\centering
\includegraphics[width=\linewidth]{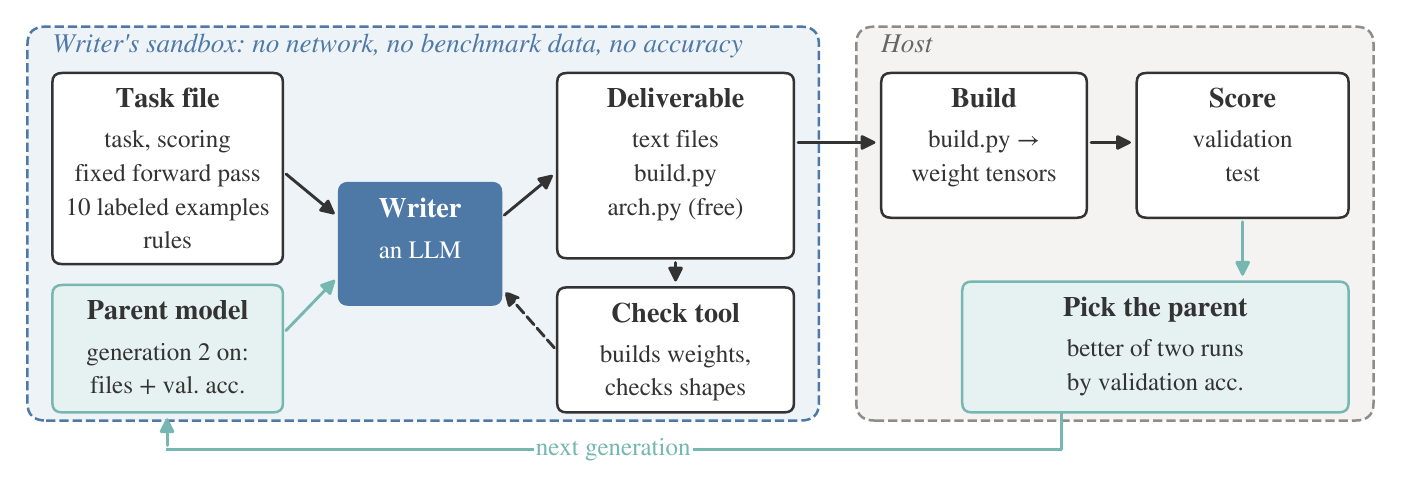}
\caption{\textbf{The writer sees only ten ARC examples and, from generation 2, its parent model and its parent's validation accuracy; only the host scores benchmark questions; of all scores, only the parent's validation accuracy passes to the next generation.} On the left, the writer produces its deliverable inside the sandbox. On the right, the host builds and scores each run.}
\label{fig:protocol}
\end{figure}

In this paper, weight-space programming (WSP) denotes the setting in which a large language model (LLM) writes the weights of a target model by hand, with no gradient training and no fitting on data (\cref{fig:protocol}). A target model is a forward pass made only of fixed tensor operations. \textit{Parameters} counts each non-zero entry of a dense tensor and, for a sparse tensor, each stored value together with its indices; zero entries of dense tensors are stored but not counted. Where a table or figure says \textit{non-zero values}, it counts non-zero entries only, without indices, and a same-sized Transformer is one with that number of parameters. A neural network such as a Transformer is one such target model, and the LLM-designed models are another. We compare models produced by WSP against models trained by gradient descent (GD) using Adam. As in the main text, examples and numbers for a single model come from Claude Opus 5.5 and its first-generation models unless another LLM or generation is named.

\paragraph{Writers and sandbox.}
The writers are Claude Opus 5.5 at high reasoning effort, GPT-6.1 Sol, and GPT-6 Astra, along with Claude Fable 5.2 where stated. Each writer works with file, search, and shell tools in an isolated sandbox without network access other than to its model. The sandbox contains no project files and no benchmark files except the examples that a task allows, and a check before each run verifies that no disallowed data file or network connection is present.

\paragraph{Task and deliverable.}
At the start of a run, the writer receives a task file with the task description, the scoring rule, example items allowed by the task, and the protocol rules. For fixed architectures, the task file also gives the forward pass exactly as evaluated and the name and shape of every weight tensor. In the sandbox, the writer can build the weights from its files, verify parameter names and shapes without running the model on benchmark data, and run the model on CPU. ARC questions are the four-option questions. On ARC, the writer sees five training questions with answers from each of ARC-Easy and ARC-Challenge and no other ARC item. For the given Transformer, our associative-memory-based model, and the LLM-designed models, the writer delivers a folder containing a word list and a build script that turns hand-written files into weight tensors, with formulas and loops allowed, plus a forward pass when designing the model. For fixed architectures, the given forward pass may not change. For a released architecture, the writer delivers an edit script that rewrites the tensors of a randomly initialized or pretrained checkpoint; across the 87 released-architecture runs from random initialization of the three writers on ARC, the median run changes 80\% of the parameters (range 35\% to 100\%), and the rest keep their random initial values. The parameter counts reported for these architectures count all of their parameters. In free design, where the writer designs the model, a static syntax check forbids Python conditionals, \texttt{while} loops, boolean operators, and conversions from tensors to Python scalars or lists, while allowing every tensor operation, including \texttt{topk}, \texttt{sort}, and \texttt{gather}. For the given Transformer, our associative-memory-based model, and the LLM-designed models, a fixed regular-expression tokenizer converts input text into word ids, keeping 128 words for each question and 48 words for each option. Scoring of released architectures uses the release's own tokenizer and the plain prompt ``\texttt{Question:\ \dots\ A.\ \dots\ D.\ \dots\ Answer:}'', reading the larger of the logits of ``\texttt{A}'' and ``\texttt{\ A}'' for each letter at the last token for decoders, Mamba-2, and RWKV-7, and at an appended mask token through the masked-language-model head for encoders. Any question on which all four options receive the same logit counts as one quarter correct for every written model, including released architectures; gradient-descent baselines and prompted pretrained models take the highest logit.

\paragraph{Given Transformer.}
The given Transformer has two pre-norm layers, four attention heads, an MLP width twice the model width, GELU activations, and learned absolute position embeddings. Each option forms one sequence made of the 128 question word ids followed by the 48 option word ids; mean pooling over non-padding tokens and a dot product with one score vector gives the option logit. In the main comparison (\cref{tab:main_results}), Claude Opus 5.5 writes widths 64, 8, and 24, and GPT-6.1 Sol and GPT-6 Astra write width 64, with additional widths written for the speed comparison. The LLM writes every weight; there are no starting weights.

\paragraph{Lineages.}
A lineage runs generations of the same task, where each generation has two runs. Generation 1 has no parent. From generation 2 on, a valid run is one whose deliverable builds and passes the checks above, and each run starts from the parent, the valid run of the previous generation with the higher accuracy on the 862 ARC validation questions (never test questions). The writer receives the parent's folder with its notes and that single validation accuracy, and nothing about which questions it got right or wrong. A lineage stops when a generation does not raise the best validation accuracy of the earlier generations, or after generation 10. The reported model of a generation is its best-validation run; the reported model of a lineage is its last generation, the one at which the lineage stopped because it did not raise the best validation accuracy, and \cref{tab:lineage_generations} gives every generation.

\begin{table}[t]
\centering
\small
\setlength{\tabcolsep}{4pt}
\resizebox{\linewidth}{!}{%
\begin{tabular}{llrrrrrrrr}
\toprule
 & & 1 & 2 & 3 & 4 & 5 & 6 & 7 & 8 \\
\midrule
Claude Opus 5.5, LLM-designed & validation & 50.7 & 52.1 & 53.5 & 53.7 & 53.0 &  &  &  \\
 & test & 50.8 & 52.2 & 53.4 & 53.3 & 54.4 &  &  &  \\
\addlinespace
Claude Opus 5.5, Associative-memory-based (ours) & validation & 47.7 & 48.9 & 51.8 & 53.4 & 53.4 &  &  &  \\
 & test & 49.6 & 51.5 & 52.1 & 52.5 & 52.4 &  &  &  \\
\addlinespace
GPT-6.1 Sol, LLM-designed & validation & 42.0 & 44.1 & 45.0 & 45.2 & 45.8 & 46.0 & 45.6 &  \\
 & test & 44.6 & 45.6 & 45.6 & 46.1 & 46.6 & 46.9 & 47.4 &  \\
\addlinespace
GPT-6.1 Sol, Associative-memory-based (ours) & validation & 40.4 & 41.9 & 42.6 & 44.1 & 43.2 &  &  &  \\
 & test & 41.2 & 43.3 & 44.1 & 45.2 & 45.6 &  &  &  \\
\addlinespace
GPT-6 Astra, LLM-designed & validation & 42.1 & 43.5 & 44.8 & 45.2 & 45.2 &  &  &  \\
 & test & 42.7 & 44.2 & 45.1 & 46.2 & 46.2 &  &  &  \\
\addlinespace
GPT-6 Astra, Associative-memory-based (ours) & validation & 42.1 & 44.4 & 45.2 & 45.5 & 46.4 & 46.4 & 47.5 & 47.3 \\
 & test & 42.5 & 44.4 & 46.1 & 46.3 & 46.9 & 47.3 & 47.3 & 47.4 \\
\bottomrule
\end{tabular}}
\caption{\textbf{Across all six ARC lineages, accuracy rises over the first generations and then levels off.} Entries report ARC validation and test accuracy (\%) of the best-validation run of each generation, which is the parent of the next.}
\label{tab:lineage_generations}
\end{table}

\paragraph{Rules.}
The protocol allows no gradient training or statistical fitting of any kind on data. During a run, the writer never sees accuracy on ARC questions other than its parent's single validation accuracy. The writer can run the model on questions it makes up and see its accuracy on them, and writers sometimes use such practice questions to track accuracy; in four runs of Claude Opus 5.5, one of them a written Transformer, the writer compared a few settings it chose on its own practice questions. On ARC data, the only choice made from measured accuracy is selecting the parent of the next generation by validation accuracy; no choice uses the test set. Formulas and code that generate weights from text files that the writer wrote by hand are allowed, whereas computing weights from dataset items is not.

\paragraph{Stopping.}
A writing run ends when the writer reports the deliverable finished or reaches a fixed budget of output tokens or wall-clock time.

\paragraph{Scoring and accounting.}
Scoring never takes place inside the writer's sandbox. After a run ends, the deliverable is copied out, built, and evaluated in a separate sandbox with no network access on the 3,530 ARC test questions, which the writer never saw. On ARC, the memory limit given to the writer is 2 GB for models other than released architectures, and scoring enforces 6 GB and 20 minutes (16 GB for released architectures). In the size comparisons of \cref{fig:teaser} and the abstract, the released architecture that Claude Opus 5.5 wrote best by validation accuracy is Qwen3-1.7B (1.72B parameters, 48.7\% test accuracy); the 680$\times$ ratio divides that count by the 2.53M non-zero values of the associative-memory-based model, and the 1{,}800$\times$ ratio divides it by the 0.94M parameters of the LLM-designed model. Reported cost is the API-price equivalent in USD of the tokens recorded during the run, including thinking tokens. For runs stopped by the time limit, the cost covers only tokens recorded before the stop.

\paragraph{Gradient descent.}
On ARC, released architectures trained from random initialization use AdamW with no weight decay, learning rates $10^{-4}$, $3\times 10^{-4}$, and $10^{-3}$, batch size 16, three epochs, and one seed, keeping the learning rate and half-epoch with the best validation accuracy. Small networks use Adam with learning rates $3\times 10^{-3}$, $10^{-3}$, and $3\times 10^{-4}$, batch size 32, 3,148 steps, and one seed, keeping the step and learning rate with the best validation accuracy. The gradient-descent rows of \cref{tab:main_results} report the architecture, size, or family with the best validation accuracy. Fine-tuning uses the instruction-tuned releases with their chat prompt where one exists, applying low-rank adaptation (LoRA, rank 16) for Transformers of at least 0.5B parameters and full fine-tuning otherwise.

\FloatBarrier

\section{Released models}\label{app:released}

\cref{tab:released_models} lists the 15 released architectures by family, type, and size, marking where each is used. Each is written from random initialization and has its pretrained base checkpoint edited by WSP and its released checkpoint fine-tuned on ARC as a reference. The nine decoders are prompted without training (\cref{tab:pretrained_open}), and all fifteen architectures are trained by gradient descent from random initialization on the ARC training set. \Cref{tab:released_wsp} reports the ARC test accuracy of each released architecture written from random initialization under each writer.

\begin{table}[t]
\centering
\small
\setlength{\tabcolsep}{4pt}
\resizebox{\linewidth}{!}{%
\begin{tabular}{lllccccc}
\toprule
 & & & \multicolumn{2}{c}{WSP} & & & \\
\cmidrule(lr){4-5}
Model & Type & Parameters & Random init. & Pretrained & Fine-tuned & Prompted & GD \\
\midrule
SmolLM2-135M~\citep{allal2025smollm2} & Decoder & 135M & \cmark & \cmark & \cmark$^\dagger$ & \cmark & \cmark \\
SmolLM2-360M~\citep{allal2025smollm2} & Decoder & 362M & \cmark & \cmark & \cmark$^\dagger$ & \cmark & \cmark \\
Qwen3-0.6B~\citep{yang2025qwen3} & Decoder & 596M & \cmark & \cmark & \cmark$^\dagger$ & \cmark & \cmark \\
Qwen3-1.7B~\citep{yang2025qwen3} & Decoder & 1.7B & \cmark & \cmark & \cmark$^\dagger$ & \cmark & \cmark \\
Qwen3.5-0.8B~\citep{qwenteam2026qwen35} & Hybrid decoder & 752M & \cmark & \cmark & \cmark$^\dagger$ & \cmark & \cmark \\
Qwen3.5-2B~\citep{qwenteam2026qwen35} & Hybrid decoder & 1.9B & \cmark & \cmark & \cmark$^\dagger$ & \cmark & \cmark \\
Gemma 3 270M~\citep{gemmateam2025gemma3} & Decoder & 268M & \cmark & \cmark & \cmark$^\dagger$ & \cmark & \cmark \\
Gemma 3 1B~\citep{gemmateam2025gemma3} & Decoder & 1.0B & \cmark & \cmark & \cmark$^\dagger$ & \cmark & \cmark \\
Gemma 2 2B~\citep{gemmateam2024gemma2} & Decoder & 2.6B & \cmark & \cmark & \cmark$^\dagger$ & \cmark & \cmark \\
ModernBERT-base~\citep{warner2025modernbert} & Encoder & 150M & \cmark & \cmark & \cmark & -- & \cmark \\
ModernBERT-large~\citep{warner2025modernbert} & Encoder & 396M & \cmark & \cmark & \cmark & -- & \cmark \\
UltraBERT-150M~\citep{park2026ultrabert} & Encoder & 190M & \cmark & \cmark & \cmark & -- & \cmark \\
UltraBERT-380M~\citep{park2026ultrabert} & Encoder & 430M & \cmark & \cmark & \cmark & -- & \cmark \\
Mamba-2 130M~\citep{dao2024mamba2} & State-space model & 129M & \cmark & \cmark & \cmark & -- & \cmark \\
RWKV-7 0.1B~\citep{peng2025rwkv7} & RNN & 191M & \cmark & \cmark & \cmark & -- & \cmark \\
\bottomrule
\end{tabular}}
\caption{\textbf{Fifteen released architectures span Transformer decoders and encoders, a state-space model, and an RNN.} Checkmarks and dashes show where each architecture is used across WSP from random initialization or a pretrained base checkpoint, ARC fine-tuning, prompting without training, and gradient descent (GD) from random initialization. Here $^\dagger$~marks post-trained, instruction-tuned checkpoints.}
\label{tab:released_models}
\end{table}

\begin{table}[t]
\centering
\small
\setlength{\tabcolsep}{4pt}
\begin{tabular}{lrrrrr}
\toprule
 & & \multicolumn{3}{c}{Weight-space programming} & \\
\cmidrule(lr){3-5}
Architecture & Parameters & Claude Opus 5.5 & GPT-6.1 Sol & GPT-6 Astra & GD \\
\midrule
Mamba-2 130M & 129M & 42.1 (2) & 34.7 (2) & 35.9 (2) & 24.6 \\
SmolLM2-135M & 135M & 44.7 (2) & 34.3 (2) & 32.8 (1) & 25.9 \\
ModernBERT-base & 150M & 41.9 (2) & 35.4 (2) & 34.7 (2) & 26.9 \\
UltraBERT-150M & 190M & 42.7 (2) & 33.3 (2) & 34.3 (2) & 25.2 \\
RWKV-7 0.1B & 191M & 42.1 (2) & 33.2 (2) & 35.9 (2) & 25.2 \\
Gemma 3 270M & 268M & 43.7 (2) & 35.3 (2) & 34.2 (2) & 25.2 \\
SmolLM2-360M & 362M & 40.3 (2) & 35.8 (2) & 33.5 (2) & 26.6 \\
ModernBERT-large & 396M & 44.2 (2) & 34.7 (2) & 34.2 (1) & 24.2 \\
UltraBERT-380M & 430M & 41.6 (2) & 37.3 (2) & 34.6 (2) & 24.2 \\
Qwen3-0.6B & 596M & 46.3 (2) & 35.7 (2) & 35.6 (2) & 24.3 \\
Qwen3.5-0.8B & 752M & 42.1 (2) & 35.3 (2) & 35.0 (2) & 26.7 \\
Gemma 3 1B & 1.0B & 37.0 (2) & 34.4 (2) & 34.6 (2) & 23.2 \\
Qwen3-1.7B & 1.7B & 48.7 (2) & 34.5 (2) & 30.6 (1) & 25.3 \\
Qwen3.5-2B & 1.9B & 43.3 (2) & 36.0 (2) & 34.9 (2) & 24.4 \\
Gemma 2 2B & 2.6B & 45.1 (2) & 35.0 (2) & 34.6 (2) & 25.2 \\
\bottomrule
\end{tabular}
\caption{\textbf{No released architecture written from random initialization reaches the LLM-designed models under any writer.} Columns report mean ARC test accuracy (\%), with run counts in parentheses. The GD column reports the same architecture trained by gradient descent on the 3,358 ARC training questions.}
\label{tab:released_wsp}
\end{table}

\FloatBarrier

\section{What the LLMs built}\label{app:models}

In the free-design setting, each LLM delivers a word list, a build script to compile hand-written text files into weight tensors, and an architecture file to run a tensor forward pass that is verified by a syntactic check against a list of forbidden Python control-flow syntax. We summarize the written line counts, vocabulary sizes, compiled tensor counts, nominal element totals, parameters, and ARC test accuracies of the three final-generation models in \cref{tab:model_inventory}.

\begin{table}[t]
\centering
\small
\setlength{\tabcolsep}{4pt}
\resizebox{\linewidth}{!}{%
\begin{tabular}{lrrrrrrr}
\toprule
LLM & Gen. & Written lines & Vocabulary & Tensors & Elements & Parameters & ARC Acc. \\
\midrule
Claude Opus 5.5 & 5 & 10{,}897 & 411{,}266 & 155 & 112M & 3{,}349{,}538 & 54.4 \\
GPT-6.1 Sol & 7 & 10{,}333 & 70{,}415 & 314 & 345M & 766{,}132 & 47.4 \\
GPT-6 Astra & 5 & 8{,}596 & 25{,}351 & 214 & 215M & 546{,}323 & 46.2 \\
\bottomrule
\end{tabular}}
\caption{\textbf{Final-generation LLM-designed models compile written knowledge lines into sparse tables whose elements are almost all zero.} Columns report generation, written lines, vocabulary, tensors, total elements, parameters counting non-zero entries plus the indices of sparse tensors, and ARC test accuracy.}
\label{tab:model_inventory}
\end{table}

\subsection{Forward passes}\label{app:models-equations}

\paragraph{Associative-memory-based model.}
We give the forward pass and the names and shapes of the weights; the LLM writes the word list, the text files, the build script, and every value (\cref{sec:ours}). Let $x_j = 1$ if a word of the question maps to concept $j > 0$ and $x_j = 0$ otherwise, and let $y_{cj}$ be the same indicator for option $c$. With the written concept weights $w_j$,
\begin{equation}
q_j = x_j\,w_j, \qquad o_{cj} = y_{cj}\,(1 - x_j)\,w_j,
\label{eq:assoc-bags}
\end{equation}
so a repeated word counts once and an option keeps only the concepts the question lacks. The forward pass takes the $J = 16$ largest entries of $q$; for a question with fewer than 16 concepts the remaining entries are zero, and every pair that contains one adds nothing to \cref{eq:assoc}. Row $h$ of the pair table is stored as $R$ concept ids and $R$ weights, and $t_h$ is the vector over concepts that adds each weight at its id. In Claude Opus 5.5's build script, pairs whose addresses coincide are merged by adding their weights concept by concept, and each row then keeps the $R$ entries with the largest absolute weight. The prime is $p = 1{,}000{,}003$; $H$, $R$, and $\mu$ are written by the LLM ($H = 2^{20}$, $R = 40$, $\mu = 1.5$ in the first-generation model of Claude Opus 5.5, and $H = 2^{22}$, $R = 16$, $\mu = 1.5$ in its fifth generation).

\paragraph{LLM-designed model of Claude Opus 5.5, first generation.}
Let $k_i$ be the 0/1 concept vector of fact $i$ and $\nu_i = \max\bigl(1, \sum_j k_{ij} w_j\bigr)^{-0.35}$ its row norm, with $w_j = \ln(1 + N/\mathrm{df}_j)$ over the $N = 4{,}288$ facts. The support of fact $i$ for option $c$ given a question vector $u$ is
\begin{equation}
s_{ci}(u) = \frac{\nu_i\,(k_i^\top u)\,(k_i^\top o_c)\,\delta_{ci}}{(\mathbf{1}^\top o_c + 1)^{0.6}},
\label{eq:designed-support}
\end{equation}
where $\delta_{ci}$ is a word-order factor that differs from 1 only when the question and option name a change between states of matter or forms of energy (for example \emph{solid} into \emph{liquid}). With $s_c^{(r)}(u)$ the $r$-th largest support over facts, and $s^{\mathrm{win}}_c$ and $s^{\mathrm{par}}_c$ the best support, without $\delta_{ci}$, over windows of four consecutive concepts and over facts joined with their neighbors, one pass scores
\begin{equation}
T_c(u) = s_c^{(1)}(u) + 0.15 \sum_{r=2}^{5} s_c^{(r)}(u) + 0.5\, s^{\mathrm{win}}_c(u) + 0.3\, s^{\mathrm{par}}_c(u).
\label{eq:designed-pass}
\end{equation}
The second pass adds to $q$ the concepts of the three facts that best match the question, $q' = q + w \odot \eta$, where $\eta$ weights the concepts of those facts by their match and is zero on the concepts of the question and the options. The option score is
\begin{equation}
\mathrm{score}_c = \pm\,(1 - 0.1\,a_c)\,\bigl(T_c(q) + T_c(q')\bigr) + 10^{-4}\,n_c,
\label{eq:designed-score}
\end{equation}
where $a_c = 1$ if option $c$ contains \emph{only}, \emph{always}, or \emph{never}, $n_c$ is the number of words of option $c$, and the sign is negative for a negated question, namely one with \emph{except}, with \emph{least} before \emph{likely} or \emph{probably}, or with a negator such as \emph{not} within seven words after the question word. The question vector $q$ weights its last 14 words twice, adds categories of its concepts from a written is-a list at 0.3, and scales each concept by how unevenly it supports the options. A route through the categories of option concepts is added with weight $e^{-\mathbf{1}^\top o_c}$, which matters only for options whose concepts the facts do not contain. \Cref{eq:designed} keeps only $s_c^{(1)}$ without $\nu_i$, $\delta_{ci}$, and the option-length scale.

\subsection{First-generation models of the three LLMs}\label{app:models-first}

\begin{figure}[t]
\centering
\captionsetup[subfigure]{justification=centering}
  \begin{subfigure}[b]{0.325\linewidth}
    \centering
    \includegraphics[width=\linewidth]{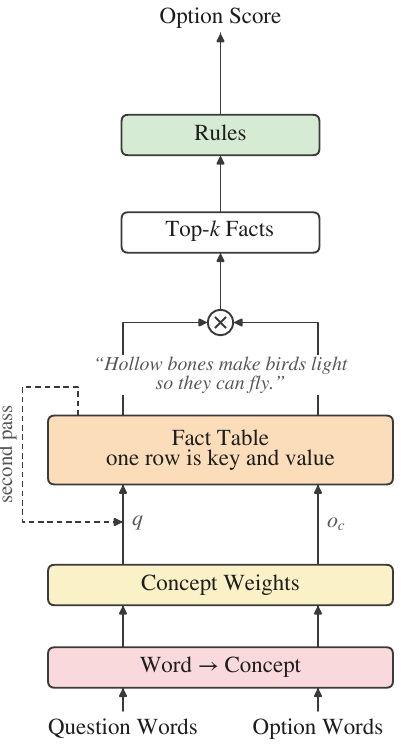}
    \subcaption{Claude Opus 5.5}\label{fig:designed-opus}
  \end{subfigure}\hfill
  \begin{subfigure}[b]{0.325\linewidth}
    \centering
    \includegraphics[width=\linewidth]{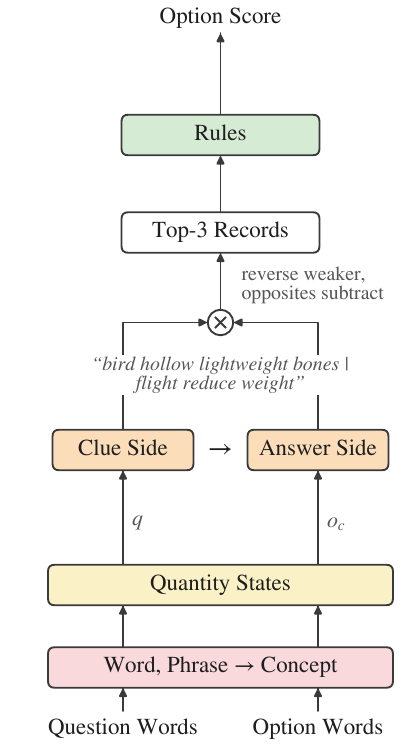}
    \subcaption{GPT-6.1 Sol}\label{fig:designed-sol}
  \end{subfigure}\hfill
  \begin{subfigure}[b]{0.325\linewidth}
    \centering
    \includegraphics[width=\linewidth]{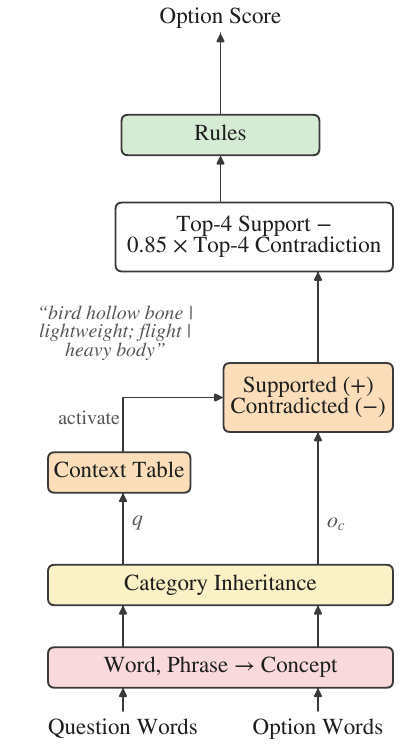}
    \subcaption{GPT-6 Astra}\label{fig:designed-astra}
  \end{subfigure}
  \caption{\textbf{All three LLMs build the same kind of first-generation model and differ in how a written row is split.} Each panel is read from bottom to top and quotes one line that its LLM wrote. Colored blocks hold weights written by the LLM, and white blocks are fixed operations.}
  \label{fig:designed_three}
\end{figure}

We compare the first-generation LLM-designed models in \cref{fig:designed_three}, using each LLM's best-validation run. With no training or search over weights, all three map question and option words to concepts and compile hand-written lines into sparse tables. Each model scores options against the best few rows and adds hand-written rule modules such as negation, comparisons, and arithmetic. They differ in how they split a written row. Claude Opus 5.5 uses one plain sentence as both key and value. GPT-6.1 Sol splits each record into a directed clue and answer, while GPT-6 Astra lists question concepts, supported answers, and contradicted answers.

\subsection{Final-generation models}\label{app:models-final}

In generation 5, Claude Opus 5.5 writes plain sentences with a single science fact per line. For example, its knowledge files contain the sentence \wfact{An advantage of solar energy is that it is renewable and produces no air pollution; a disadvantage is that it does not work at night or on cloudy days and solar panels are expensive.} When compiling the deliverable, the build step maps every vocabulary token to a concept identifier through a lookup table of 411,268 entries (411,266 words and two reserved ids). It then stores fact-by-concept tables as dense index arrays and values covering whole facts, windows of four consecutive concepts, facts paired with their neighbors, and single clauses.

In generation 7, GPT-6.1 Sol organizes domain knowledge as rules of the form question cues followed by answer cues, for example \texttt{force measurement unit} then \texttt{newton}. During compilation, the build step turns these written rules into two sparse matrices spanning 10,382 rules by 9,626 concepts, with 40,699 non-zero entries on the cue side and 45,199 on the answer side.

In generation 5, GPT-6 Astra writes structured lines specifying question concepts, supported answers, and contradicted answers, for example \texttt{balanced\_force}, then \texttt{net\_force zero; constant velocity; no acceleration}, then \texttt{always motionless}. From these written lines, the build step forms two sparse matrices over 5,518 concepts, a concept-by-context matrix across 11,410 question contexts with 29,681 non-zero entries and a concept-by-assertion matrix across 22,398 answer assertions with 53,437 non-zero entries.

Across all three lineages, almost every element of the compiled tables is zero, and the forward pass scores options by question-option match alongside small rule modules.

\FloatBarrier

\section{Controls on the written models}\label{app:controls}

\paragraph{Test-set overlap.}
The text written by each LLM overlaps the ARC test set about as often as comparable questions that the LLM did not see, and deleting the overlapping lines barely changes test accuracy (\cref{tab:contamination}). Across all three lineages, overlap on ARC test matches the low rate on the ARC training questions that the LLM never saw during the run and on external science benchmarks. In this test, an item is the question stem followed by its correct option in lowercase word tokens, and a written artifact is every line of every file in the model folder. An item is flagged if any of four rules fires. Under the first rule, any word 11-gram of an ARC-Easy item, 12-gram of an ARC-Challenge item, or 13-gram of another item occurs in one line~\citep{brown2020language}. Under the second rule, at least 70\% of the item's 8-grams occur in the artifact~\citep{chowdhery2022palm}. Under the third rule, the longest token span shared with one line covers at least half of the item~\citep{singh2024evaluation}; the 0.5 threshold is ours. The fourth rule is ours, with content words defined as tokens other than stopwords and digits. For items with at least five content words, it fires when one line holds at least 80\% of the item's content words and every content word of its correct option. The deletion test follows \citet{gunasekar2023textbooks}. When Claude Opus 5.5 guesses a masked option from an item stem, exact match is 15.3\% on ARC test. The same probe reaches 16.3\% on ARC training items, so Claude Opus 5.5 recalls masked ARC test options no more often than ARC training options.

\begin{table}[t]
\centering
\small
\setlength{\tabcolsep}{3.5pt}
\resizebox{\textwidth}{!}{%
\begin{tabular}{lrrrrrrrrrr}
\toprule
 & & \multicolumn{4}{c}{Items overlapping a written line (\%)} & & \multicolumn{4}{c}{ARC test accuracy (\%)} \\
\cmidrule(lr){3-6}\cmidrule(lr){8-11}
 & & ARC & ARC train & OpenBookQA & SciQ & Lines & \multicolumn{2}{c}{All items} & \multicolumn{2}{c}{Unflagged items} \\
\cmidrule(lr){8-9}\cmidrule(lr){10-11}
LLM & Generation & test & (unseen) & test & test & deleted & Original & Deleted & Original & Deleted \\
\midrule
Claude Opus 5.5 & 1 & 1.6 & 1.5 & 0.0 & 2.7 & 51 & 50.8 & 50.6 & 50.2 & 50.2 \\
 & 5 & 2.1 & 2.0 & 0.0 & 4.0 & 68 & 54.4 & 54.3 & 53.7 & 53.7 \\
\midrule
GPT-6.1 Sol & 1 & 0.5 & 0.6 & 0.0 & 0.6 & 16 & 44.6 & 44.2 & 44.3 & 44.1 \\
 & 7 & 1.0 & 0.8 & 0.0 & 1.4 & 34 & 47.4 & 46.8 & 47.0 & 46.6 \\
\midrule
GPT-6 Astra & 1 & 0.2 & 0.2 & 0.0 & 0.2 & 7 & 42.7 & 42.7 & 42.6 & 42.7 \\
 & 5 & 0.5 & 0.5 & 0.0 & 0.7 & 17 & 46.2 & 46.2 & 45.9 & 46.0 \\
\bottomrule
\end{tabular}}
\caption{\textbf{Written lines rarely overlap ARC test items, and deleting every overlapping line barely changes accuracy.} An item overlaps when it matches a line the LLM wrote under any of four overlap rules (three published, one ours; \cref{app:controls}). Deleted gives test accuracy after deleting every flagged line and rebuilding.}
\label{tab:contamination}
\end{table}

\paragraph{No practice questions.}
Forbidding the LLM from writing practice questions of its own leaves test accuracy about the same as in the unrestricted runs. Because standard runs let an LLM write practice questions to check its forward pass, we test Claude Opus 5.5 under a rule that forbids question files and allows one placeholder question for checking the forward pass. Both restricted runs obey this rule, and the two runs average about one point below the unrestricted runs (\cref{tab:controls}).

\paragraph{Line-removal test.}
In the line-removal test of \cref{tab:line_ablation}, the weights column is the median number of values at fixed tensor indices that change by over 1\%, which is defined when removing a line keeps column indices fixed, and the answers column is the mean. Accuracy in that table takes the highest-scoring option for every question, whereas \cref{tab:main_results} counts a question on which all four options tie as one quarter correct, so the two differ by up to 0.3 points.

\begin{table}[t]
\centering
\small
\setlength{\tabcolsep}{4pt}
\resizebox{\linewidth}{!}{%
\begin{tabular}{lrrrrrrr}
\toprule
 & Non-zero & Weights & Answers & \multicolumn{2}{c}{Share a word with the line (\%)} & \multicolumn{2}{c}{Test accuracy (\%)} \\
\cmidrule(lr){5-6}\cmidrule(lr){7-8}
Model & values (M) & changed & changed & changed answers & all questions & before & after \\
\midrule
\multicolumn{8}{l}{\textit{GPT-6.1 Sol}} \\
LLM-designed, run 1 & 0.13 & -- & 1.0 & 96 & 7 & 42.9 & 42.9 \\
LLM-designed, run 2 & 0.11 & -- & 1.1 & 75 & 13 & 44.6 & 44.6 \\
Associative-memory-based (ours), run 1 & 0.07 & 19 & 1.6 & 54 & 8 & 42.4 & 42.4 \\
Associative-memory-based (ours), run 2 & 0.12 & 168 & 5.7 & 41 & 7 & 41.3 & 41.3 \\
Written Transformer, run 2 & 0.17 & 281 & 10.8 & 97 & 7 & 34.4 & 34.3 \\
\midrule
\multicolumn{8}{l}{\textit{GPT-6 Astra}} \\
LLM-designed, run 1 & 0.16 & -- & 1.8 & 94 & 9 & 42.7 & 42.7 \\
LLM-designed, run 2 & 0.09 & 58 & 0.5 & 100 & 7 & 41.4 & 41.4 \\
Associative-memory-based (ours), run 1 & 0.08 & 10 & 0.4 & 100 & 6 & 43.5 & 43.5 \\
Associative-memory-based (ours), run 2 & 0.09 & 103 & 8.7 & 39 & 6 & 42.6 & 42.6 \\
Written Transformer, run 1 & 0.31 & 2{,}223 & 20.3 & 90 & 10 & 32.6 & 32.6 \\
\bottomrule
\end{tabular}}
\caption{\textbf{In the first-generation models of GPT-6.1 Sol and GPT-6 Astra, a removed line also changes fewer answers in a table model than in a written Transformer.} The test and columns match \cref{tab:line_ablation}, and -- in the weights column marks a model that renumbers its table columns when a line is removed, so its changed weights cannot be counted.}
\label{tab:line_ablation_writers}
\end{table}

\paragraph{BM25 for all three LLMs.}
In \cref{tab:bm25_writers}, stemmed BM25 searches the text files, the tables, and Python string literals, because GPT-6.1 Sol and GPT-6 Astra often write knowledge as Python strings, whereas \cref{tab:bm25} searches the text files alone. Written two-layer Transformers score below BM25 over their text in 20 of 23 runs, with 12 of 12 for Claude Opus 5.5, 4 of 6 for GPT-6.1 Sol, and 4 of 5 for GPT-6 Astra. LLM-designed and associative-memory-based table models score above it in 21 of 24 runs, at 8 of 8, 7 of 8, and 6 of 8 across the three LLMs. In written released architectures, 48 of the 54 GPT-6.1 Sol and GPT-6 Astra runs score above BM25 over their text, whose BM25 means are 31.1\% and 31.5\%, whereas 24 of Claude Opus 5.5's 30 score below it.

\begin{table}[t]
\centering
\small
\setlength{\tabcolsep}{4pt}
\resizebox{\linewidth}{!}{%
\begin{tabular}{llrrr}
\toprule
LLM & Model & Written model & BM25 over its text & Difference \\
\midrule
Claude Opus 5.5 & Written two-layer Transformer & 36.1 & 43.9 & $-$7.8 \\
 & Written two-layer Transformer, speed widths & 39.9 & 45.4 & $-$5.5 \\
 & Written released architecture & 43.0 & 45.4 & $-$2.3 \\
 & LLM-designed, generation 1 & 50.7 & 46.8 & $+$3.9 \\
 & LLM-designed, generation 5 & 54.1 & 48.4 & $+$5.7 \\
 & Associative-memory-based (ours), generation 1 & 49.9 & 48.4 & $+$1.5 \\
 & Associative-memory-based (ours), generation 5 & 52.6 & 49.1 & $+$3.5 \\
\midrule
GPT-6.1 Sol & Written two-layer Transformer & 36.3 & 35.2 & $+$1.1 \\
 & Written two-layer Transformer, speed widths & 33.7 & 38.2 & $-$4.5 \\
 & Written released architecture & 34.9 & 31.1 & $+$3.8 \\
 & LLM-designed, generation 1 & 43.7 & 42.0 & $+$1.7 \\
 & LLM-designed, generation 7 & 47.3 & 46.2 & $+$1.2 \\
 & Associative-memory-based (ours), generation 1 & 41.8 & 40.6 & $+$1.2 \\
 & Associative-memory-based (ours), generation 5 & 45.8 & 42.1 & $+$3.8 \\
\midrule
GPT-6 Astra & Written two-layer Transformer & 32.6 & 40.7 & $-$8.1 \\
 & Written two-layer Transformer, speed widths & 32.8 & 34.7 & $-$1.9 \\
 & Written released architecture & 34.6 & 31.5 & $+$3.1 \\
 & LLM-designed, generation 1 & 42.1 & 39.2 & $+$2.9 \\
 & LLM-designed, generation 5 & 46.3 & 42.6 & $+$3.7 \\
 & Associative-memory-based (ours), generation 1 & 43.0 & 43.1 & $-$0.1 \\
 & Associative-memory-based (ours), generation 8 & 47.4 & 47.2 & $+$0.2 \\
\bottomrule
\end{tabular}}
\caption{\textbf{For every LLM, written Transformers mostly score below a plain search over their own text, and table models mostly score above it.} Columns report mean ARC test accuracy (\%) over the runs of each kind. Here \textit{speed widths} are additional Transformer widths from the speed comparison.}
\label{tab:bm25_writers}
\end{table}

\paragraph{Where writing facts stops.}
A written table model holds only explicit facts and rules that the writer states in text. In the LLM-designed model, 45\% of all 1,165 ARC-Challenge test questions have a fact that shares a concept with the question and a concept new to the question with the correct option, while a wrong option's best fact scores higher, so the reader's word matching limits accuracy as well as what the writer stated.

\FloatBarrier

\section{Inference efficiency in detail}\label{app:speed}

\Cref{tab:speed_fastest} gives the GPU time of the same-sized Transformer under three input treatments, with every question padded to 176 positions, each batch cut to its real length, or questions sorted by length and then cut. The main text compares with the last, the fastest. \Cref{tab:gpu_rewrite} gives the two LLM-designed models that we asked Claude Opus 5.5 to rewrite for a GPU, with two rewrites of each. The two generation-5 rewrites differ in time because the slower one batches both passes into intermediate tensors twice as large (its summed kernel time is 6.1 s against 1.7 s), even though it launches fewer kernels.
\begin{table}[t]
\centering
\small
\setlength{\tabcolsep}{5pt}
\begin{tabular}{lrrrr}
\toprule
 & \multicolumn{2}{c}{2.5M values} & \multicolumn{2}{c}{6.2M values} \\
\cmidrule(lr){2-3}\cmidrule(lr){4-5}
 & Time (ms) & Slower & Time (ms) & Slower \\
\midrule
Associative-memory-based (ours) & 18 & -- & 18 & -- \\
Transformer, inputs padded & 678 & 37$\times$ & 1530 & 85$\times$ \\
Transformer, inputs cut to length & 273 & 15$\times$ & 633 & 35$\times$ \\
Transformer, sorted and cut & 134 & 7$\times$ & 302 & 17$\times$ \\
\bottomrule
\end{tabular}
\caption{\textbf{The associative-memory-based model answers the test set in the same time at both sizes, 7 to 17 times faster than the fastest same-sized Transformer.} Columns report CUDA-graph time on one RTX 3090 for all 3,530 ARC test questions in batches of 64, matching each two-layer Transformer to our model's non-zero values. \textit{Slower} is relative to our model (--).}
\label{tab:speed_fastest}
\end{table}

\begin{table}[t]
\centering
\small
\setlength{\tabcolsep}{4pt}
\resizebox{\linewidth}{!}{%
\begin{tabular}{lrrrrl}
\toprule
 & Values & ARC acc. & GPU time (s) & Transformer (s) & vs.\ Transformer \\
\midrule
LLM-designed, generation 1 & 0.57M & 50.8 & 0.566 & 0.052 & 11$\times$ slower \\
\quad rewritten by the LLM for a GPU & 0.57M & 50.8 & 0.486 & 0.052 & 9$\times$ slower \\
\quad rewritten by the LLM for a GPU & 0.66M & 51.2 & 0.325 & 0.058 & 6$\times$ slower \\
\midrule
LLM-designed, generation 5 & 3.35M & 54.4 & 15.8 & 0.176 & 90$\times$ slower \\
\quad rewritten by the LLM for a GPU & 3.07M & 54.4 & 1.50 & 0.151 & 10$\times$ slower \\
\quad rewritten by the LLM for a GPU & 3.34M & 54.5 & 5.74 & 0.176 & 33$\times$ slower \\
\bottomrule
\end{tabular}}
\caption{\textbf{The LLM's own rewrites for a GPU stay slower than a same-sized Transformer.} Columns report non-zero values, ARC test accuracy (\%), and GPU time for the 3,530 test questions on one RTX 3090 across Claude Opus 5.5 models and same-sized two-layer Transformers. Each time is the fastest of eager execution, a CUDA graph, and \texttt{torch.compile}.}
\label{tab:gpu_rewrite}
\end{table}

\paragraph{What the architecture keeps.}
\Cref{tab:design_reasons} lists, for each way WSP differs from gradient training, what we measured and the choice it leads to in the associative-memory-based model.

\begin{table}[t]
\centering
\small
\setlength{\tabcolsep}{5pt}
\begin{tabular}{>{\raggedright\arraybackslash}p{0.28\linewidth}>{\raggedright\arraybackslash}p{0.32\linewidth}>{\raggedright\arraybackslash}p{0.32\linewidth}}
\toprule
WSP observation & Measured & Design choice (ours) \\
\midrule
Facts written one per line & all three LLMs (\cref{app:models}) & line-by-line authoring retained \\
\addlinespace
Single-line trace and local edits & 0.8--0.9 of 3{,}530 answers change per removed line & explicit concept and pair entries \\
\addlinespace
Dense overlapping keys blur facts & written Transformer trails BM25 by 8.0 points & sparse keys at hashed addresses \\
\addlinespace
Full scan over stored facts & GPU time grows from 0.57 s to 15.8 s (gen.\ 1 to 5) & fixed 120 pair reads plus one product with $A$ \\
\addlinespace
LLM GPU rewrites remain slow & 6--33$\times$ slower than same-sized Transformer & fixed forward pass replayed as CUDA graph (18 ms) \\
\bottomrule
\end{tabular}
\caption{\textbf{Each design choice in our associative-memory-based model addresses an empirical observation from WSP.} Measurements are from Claude Opus 5.5 unless noted.}
\label{tab:design_reasons}
\end{table}

\begin{figure}[t]
\centering
  \begin{subfigure}{0.49\linewidth}
    \centering
    \includegraphics[width=\linewidth]{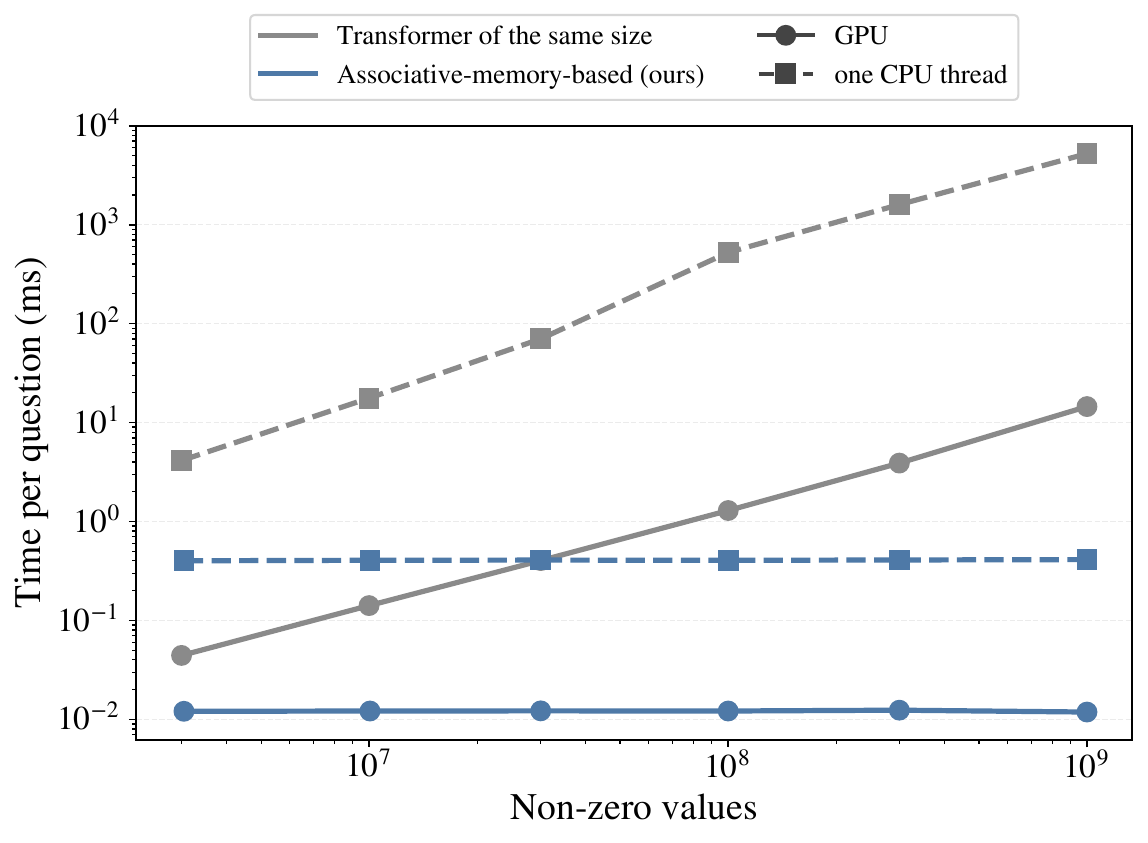}
    \subcaption{Growing the pair table}\label{fig:scale-values}
  \end{subfigure}\hfill
  \begin{subfigure}{0.49\linewidth}
    \centering
    \includegraphics[width=\linewidth]{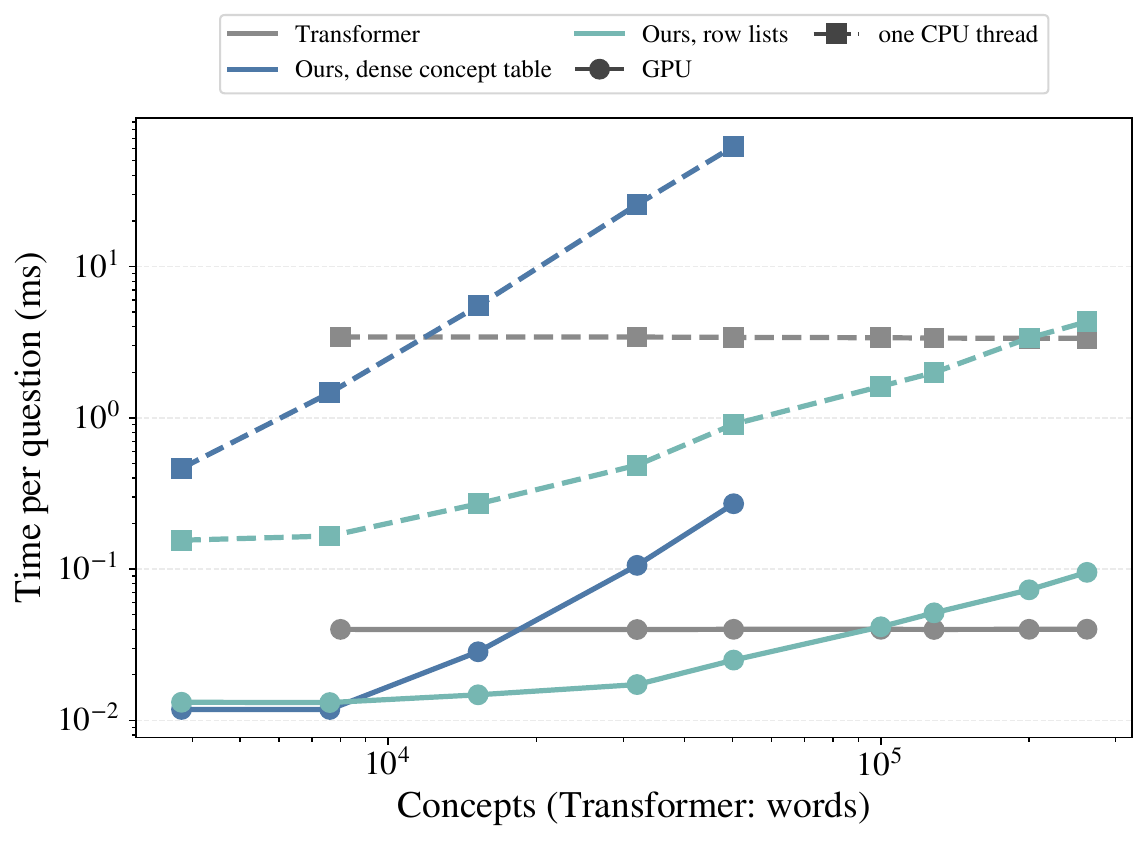}
    \subcaption{Growing the number of concepts}\label{fig:scale-concepts}
  \end{subfigure}
  \caption{\textbf{Time per question of the associative-memory-based model does not depend on how many values the pair table holds, and grows with the number of concepts.} Panel (a) varies pair-table non-zero values at 3,805 concepts, and (b) varies concepts (or Transformer vocabulary). Runs use synthetic tables and the ARC test questions on one RTX 3090 or one CPU thread.}
  \label{fig:scale}
\end{figure}

\paragraph{Scaling.}
We measure scaling with synthetic tables of the written models' shape on the real ARC test questions in eager execution. Growing the pair table from 3 million to 1 billion non-zero values raises its row count from 32,768 to 33,554,432 (\cref{fig:scale-values}). Time per question stays at 0.012 ms on the GPU and about 0.4 ms on one CPU thread. The same-sized Transformer at its fastest input goes from 0.044 ms to 14.5 ms per question on the GPU. With the concept table stored densely, time per question rises from 0.012 ms at 3,805 concepts to 0.27 ms at 50,257 concepts, and we measured the dense table up to 50,257 concepts (\cref{fig:scale-concepts}). A row-list variant reads the rows of the question's 32 strongest concepts, with each row holding up to 512 entries. It picks the same option as the dense table on at least 99.9\% of the questions across the measured dense-table sizes, and takes 0.095 ms at 262,144 concepts. Its remaining growth is linear in the number of concepts, because the forward pass still builds one vector over all concepts for each question. In panel (b), the Transformer keeps width 216 and two layers; only its vocabulary grows, from 2.52M to 57.4M parameters, and it takes 0.040 ms at every size. On the GPU, the associative-memory-based model becomes slower than this Transformer between about 15,000 and 32,000 concepts with the concept table stored densely, as in the written models, and beyond about 100,000 concepts in the row-list variant.

\paragraph{Capacity and interference.}
Rows of the pair table are addressed by a hash, so different pairs can share a row at any table size (757 rows are shared at 1,048,576 rows), which is the interference of this associative-memory-based model. Folding that pair table from 1,048,576 to 65,536 rows keeps the accuracy (\cref{tab:capacity}). With fewer rows, accuracy decreases. An edit stays local in the associative-memory-based model as it does in the LLM-designed models. Removing one written line changes two answers or fewer on average (\cref{tab:line_ablation}).

\begin{table}[t]
\centering
\small
\setlength{\tabcolsep}{4pt}
\begin{tabular}{rr}
\toprule
Pair-table rows & ARC acc. \\
\midrule
1{,}048{,}576 & 49.6 \\
262{,}144 & 49.7 \\
65{,}536 & 49.8 \\
16{,}384 & 47.4 \\
4{,}096 & 44.0 \\
\bottomrule
\end{tabular}
\caption{\textbf{Folding the pair table to one sixteenth of its rows keeps the accuracy, and folding further loses it.} Columns report pair-table row count and ARC test accuracy (\%) when the associative-memory-based model's pair table is folded to fewer rows.}
\label{tab:capacity}
\end{table}

\paragraph{Speed comparison.}
In \cref{tab:speed_compare}, an LLM-designed model is timed as the LLM wrote it, and the associative-memory-based model is timed as a CUDA graph. Between the two runs of our associative-memory-based model, the best-validation run reported there builds a dense concept table, whereas the second run builds a sparse one and takes 70 ms. The fifth-generation associative-memory-based model has 4,324 concepts. The same-sized Transformer's accuracy is the mean of its two runs, and its GPU time is at its fastest input, with questions sorted by length and each batch cut to its real length. Inputs are already encoded and on the GPU, and sorting the questions by length and restoring their order is not timed. Where the Transformer's inputs are cut to each batch's length, it uses one CUDA graph per batch shape, captured before timing, and each reported time is the sum over batches of the median of three replays. The first-generation LLM-designed model stores its tables sparsely, so \cref{tab:main_results}, which also counts the stored indices, lists it with 0.94 million parameters.

\FloatBarrier

\section{Further experiments in detail}\label{app:further}

\subsection{WSP of pretrained weights}\label{app:further-pretrained}
LLMs also edited released pretrained base checkpoints on ARC, where unedited checkpoints are scored from a plain prompt and letter logits in the same way as models written by WSP. An encoder is read at a mask token appended after ``Answer:'' through its masked-language-model head, while decoders, Mamba-2, and RWKV-7 are read at the last token. The writer could run the starting checkpoint on CPU in its sandbox, and Claude Opus 5.5 ran it on the ten example questions in 16 of its 29 runs. Editing leaves strong bases near their unedited test accuracy, lifts weak ones, and in some cases damages the base. In \cref{tab:pretrained_edit}, the checkpoints that score above 75\% stay within about one point of their unedited accuracy, while the ones near chance gain up to about twelve points. On UltraBERT-380M, editing reduces accuracy from 63.7\% to 33.1\% and 33.2\% under two of the three LLMs. Small edits and readout rewirings only adjust or route the existing checkpoint, while overwrites discard the pretrained weights.

\begin{table}[t]
\centering
\small
\setlength{\tabcolsep}{4pt}
\begin{tabular}{lrrrr}
\toprule
Pretrained checkpoint & Unedited & Claude Opus 5.5 & GPT-6.1 Sol & GPT-6 Astra \\
\midrule
Gemma 3 270M & 24.0 & 31.8 & 29.4 & 35.0 \\
Mamba-2 130M & 24.5 & 29.7 & 34.9 & 32.2 \\
Gemma 3 1B & 25.0 & 30.5 & 24.8 & 25.0 \\
SmolLM2-135M & 25.0 & 34.4 & 25.1 & 29.7 \\
RWKV-7 0.1B & 25.2 & 37.4 & 33.7 & 28.7 \\
SmolLM2-360M & 26.2 & 36.1 & 26.2 & 29.1 \\
ModernBERT-base & 26.4 & 31.5 & 34.7 & 35.2 \\
ModernBERT-large & 37.4 & 38.9 & 33.7 & 33.7 \\
UltraBERT-150M & 43.3 & 42.8 & 34.8 & 32.3 \\
UltraBERT-380M & 63.7 & 63.7 & 33.1 & 33.2 \\
Gemma 2 2B & 72.3 & 75.3 & 72.3 & 72.4 \\
Qwen3-0.6B & 75.4 & 75.4 & 75.4 & 75.4 \\
Qwen3.5-0.8B & 80.8 & 80.3 & 80.8 & 80.6 \\
Qwen3-1.7B & 87.9 & 87.9 & 87.9 & 87.9 \\
Qwen3.5-2B & 91.0 & 91.0 & 90.8 & 91.0 \\
\bottomrule
\end{tabular}
\caption{\textbf{Editing lifts checkpoints that start near chance, leaves strong checkpoints where they were, and in some cases damages a checkpoint.} ARC test accuracy of fifteen released pretrained checkpoints, sorted by unedited accuracy, unedited and after each LLM edited the weights (mean of runs).}
\label{tab:pretrained_edit}
\end{table}

\paragraph{Kinds of edit.}
Across the 87 runs, 48 change under 1\% of parameters. Of these, 30 move science-word embeddings toward hand-written definitions or add fact neurons, 11 calibrate letter read-out, 4 route an existing signal to the letters, and 3 change nothing. In 43 of the 48, accuracy stays within one point of the unedited checkpoint; the five exceptions are all by Claude Opus 5.5 and range from $-$1.3 to $+$5.6 points. Another 30 overwrite the checkpoint, 29 with a hand-written rule or fact model and one with a hand-built similarity circuit. Of these 30 overwrites, 19 are on checkpoints near chance (24.0\% to 26.4\%), and all three Claude Opus 5.5 overwrites are among them. The other 11 are GPT-6.1 Sol and GPT-6 Astra runs on the three masked-language encoders that start above chance. Two GPT-6.1 Sol runs and two GPT-6 Astra runs overwrite ModernBERT-large, which starts at 37.4\%; two GPT-6.1 Sol runs and one GPT-6 Astra run overwrite UltraBERT-150M, which starts at 43.3\%; and two GPT-6.1 Sol runs and two GPT-6 Astra runs overwrite UltraBERT-380M, which starts at 63.7\%. Each of these 11 overwrites lowers accuracy. No checkpoint at 72.3\% or above is overwritten across 28 runs, each changing at most 0.07\% of parameters. The remaining nine keep the checkpoint, rewiring late layers or routing an existing signal so that its own language-model signal selects the letter.

We also score public pretrained models without training, using a plain prompt with no examples on ARC, MMLU and MathBench and the task's ten examples in the prompt on SQuAD 1.1 and SciQ (\cref{tab:pretrained_open}). Test sets contain 3,530 items on ARC, 14,042 on MMLU, 1,121 on MathBench, 10,570 on SQuAD 1.1, and 1,000 on SciQ, and MMLU uses the first development question of ten subjects as its ten examples. With the task's ten examples in the prompt on ARC, MMLU, and MathBench (\cref{tab:pretrained_open_10shot}), the pretrained models at chance stay within 2.0 points of chance. Among the remaining base models, accuracy changes by at most 3.2 points except on Gemma 2 2B, which gains 5.2 points on ARC and 9.8 on MathBench. The instruction-tuned checkpoint and its edited version gain 3.3 to 12.5 points. On reading a passage (SQuAD 1.1), the LLM-designed model of Claude Opus 5.5, written with no training and averaged over two runs per task, is close to the smallest pretrained models. On writing out an answer from its own weights (SciQ) and on ARC, where the two-run mean is 50.7\%, it is far below the pretrained models that are above chance. On MMLU it is likewise far below them. On MathBench it scores 40.3\%, within one point of Gemma 2 2B with no examples (41.1\%) and below the other pretrained models above chance. With ten examples, Gemma 2 2B reaches 50.9\%. The MathBench models combine lookups over written facts with small rules for integer properties and one-step conversions, and perform no multi-step calculation. The edited checkpoint that \citet{takehi2026weightspace} released scores within half a point of its unedited base on ARC, MMLU and MathBench and slightly below it on SQuAD in our evaluation.

\begin{table}[t]
\centering
\small
\setlength{\tabcolsep}{4pt}
\resizebox{\linewidth}{!}{%
\begin{tabular}{lrrrrrr}
\toprule
Model & Parameters & ARC & MMLU & MathBench & SQuAD 1.1 (F1) & SciQ, written out (F1) \\
\midrule
SmolLM2-135M & 135M & 25.0 & 25.2 & 24.7 & 46.9 & 36.8 \\
Gemma 3 270M & 268M & 24.0 & 26.0 & 25.8 & 45.7 & 31.3 \\
SmolLM2-360M & 362M & 26.2 & 25.8 & 24.4 & 67.5 & 50.4 \\
Qwen3-0.6B & 596M & 75.4 & 49.7 & 58.6 & 82.5 & 52.5 \\
Qwen3.5-0.8B & 752M & 80.8 & 52.3 & 60.8 & 85.5 & 52.9 \\
Gemma 3 1B & 1.0B & 25.0 & 26.1 & 27.1 & 72.1 & 50.4 \\
Qwen3-1.7B & 1.7B & 87.9 & 60.4 & 70.1 & 88.1 & 62.9 \\
Qwen3.5-2B & 1.9B & 91.0 & 62.3 & 68.5 & 89.6 & 61.5 \\
Gemma 2 2B & 2.6B & 72.4 & 49.3 & 41.1 & 86.1 & 63.0 \\
\midrule
Qwen3.5-0.8B (instruct) & 752M & 73.5 & 46.7 & 49.2 & 83.1 & 49.4 \\
\quad edited, released by \citet{takehi2026weightspace} & 752M & 73.9 & 47.0 & 49.6 & 82.3 & 49.2 \\
\midrule
LLM-designed models, no training & -- & 50.7 & 31.0 & 40.3 & 42.3 & 19.0 \\
\bottomrule
\end{tabular}}
\caption{\textbf{An LLM-designed model written with no training is close to the smallest pretrained models on reading a passage and below the pretrained models that are above chance elsewhere.} ARC, MMLU, and MathBench report accuracy (\%, chance 25). Pretrained rows are prompted without training, and the last row averages two Claude Opus 5.5 models written separately per task.}
\label{tab:pretrained_open}
\end{table}

\begin{table}[t]
\centering
\small
\setlength{\tabcolsep}{4pt}
\resizebox{\linewidth}{!}{%
\begin{tabular}{lrrrrrrr}
\toprule
& & \multicolumn{2}{c}{ARC} & \multicolumn{2}{c}{MMLU} & \multicolumn{2}{c}{MathBench} \\
\cmidrule(lr){3-4}\cmidrule(lr){5-6}\cmidrule(lr){7-8}
Model & Parameters & none & ten & none & ten & none & ten \\
\midrule
SmolLM2-135M & 135M & 25.0 & 24.6 & 25.2 & 24.7 & 24.7 & 24.0 \\
Gemma 3 270M & 268M & 24.0 & 23.4 & 26.0 & 27.0 & 25.8 & 24.8 \\
SmolLM2-360M & 362M & 26.2 & 24.6 & 25.8 & 24.5 & 24.4 & 25.2 \\
Qwen3-0.6B & 596M & 75.4 & 78.5 & 49.7 & 50.8 & 58.6 & 61.7 \\
Qwen3.5-0.8B & 752M & 80.8 & 80.2 & 52.3 & 52.6 & 60.8 & 62.5 \\
Gemma 3 1B & 1.0B & 25.0 & 26.1 & 26.1 & 26.8 & 27.1 & 23.9 \\
Qwen3-1.7B & 1.7B & 87.9 & 88.8 & 60.4 & 61.0 & 70.1 & 70.0 \\
Qwen3.5-2B & 1.9B & 91.0 & 90.7 & 62.3 & 61.3 & 68.5 & 70.0 \\
Gemma 2 2B & 2.6B & 72.4 & 77.6 & 49.3 & 51.7 & 41.1 & 50.9 \\
\midrule
Qwen3.5-0.8B (instruct) & 752M & 73.5 & 79.3 & 46.7 & 50.9 & 49.2 & 61.6 \\
\quad edited, released by \citet{takehi2026weightspace} & 752M & 73.9 & 79.5 & 47.0 & 50.3 & 49.6 & 62.1 \\
\bottomrule
\end{tabular}}
\caption{\textbf{Ten examples in the prompt leave the pretrained models at chance there and move most others by a few points.} Columns report test accuracy (\%, chance 25) on 3,530 ARC, 14,042 MMLU, and 1,121 MathBench items, with no prompt examples in \textit{none} and the task's ten examples in \textit{ten}.}
\label{tab:pretrained_open_10shot}
\end{table}

\subsection{Other benchmarks}\label{app:further-transfer}
Models written for ARC score within about four points of their ARC accuracy on SciQ, score lower on OpenBookQA, and score near chance on MMLU (\cref{tab:transfer}).

\begin{table}[t]
\centering
\small
\setlength{\tabcolsep}{5pt}
\begin{tabular}{lrrrrr}
\toprule
 & & & & \multicolumn{2}{c}{MMLU} \\
\cmidrule(lr){5-6}
Model written for ARC & ARC & SciQ & OpenBookQA & Science & Other \\
\midrule
LLM-designed, generation 1 & 50.8 & 47.7 & 31.8 & 26.5 & 25.7 \\
LLM-designed, generation 5 & 54.4 & 50.1 & 31.7 & 25.8 & 25.9 \\
Associative-memory-based (ours), generation 1 & 49.6 & 47.7 & 30.3 & 26.4 & 25.2 \\
Associative-memory-based (ours), generation 5 & 52.4 & 50.1 & 36.1 & 27.5 & 25.9 \\
\midrule
Written two-layer Transformer, width 24 & 36.6 & 36.9 & 25.1 & 27.5 & 25.2 \\
Written two-layer Transformer, width 8 & 38.8 & 39.8 & 30.1 & 25.5 & 25.2 \\
\bottomrule
\end{tabular}
\caption{\textbf{Models written for ARC keep their accuracy on SciQ, drop on OpenBookQA, and fall to chance on MMLU.} Accuracy (\%) for Claude Opus 5.5 models run unchanged, one run per row; SciQ has 1,000 questions, OpenBookQA 500, and chance is 25\%.}
\label{tab:transfer}
\end{table}

\subsection{Text generation tasks}\label{app:further-generation}
Each task gives ten example items (10,570 test items on SQuAD 1.1 and 1,000 on SciQ), and the open-answer experiments in \cref{tab:open_tasks} use Claude Opus 5.5. For SQuAD, the LLM-designed models hold word lists (word forms, words used for one another, category lists of names, places, and diseases, and which question words ask for which kind of answer) and a fixed reading procedure, with no question-answer facts. For SciQ, all models hold a table of answers with the words that a question about them contains. They write out the stored answer of the best-matching row, and the LLM-designed model is slightly above the nearest-question lookup reference, which searches all 11,679 training questions by word overlap. The given Transformer for SQuAD reads the paragraph and the question and outputs start and end logits per position, decoded as the best span of at most 30 words; for SciQ it reads the question and the answer so far and is decoded greedily for at most five words. The trained Transformer uses Adam, three learning rates, three seeds, and the SQuAD 1.1 training items with an answer span (85,864), along with all 11,679 SciQ training questions.

\subsection{Several tasks at once}\label{app:further-tasks}
When the four kinds of items are named, a single model written by Claude Opus 5.5 reaches 45.0\% on ARC, every other multiple-choice score stays within seven points of chance, and scores are lower when no task is named (\cref{tab:general_tasks}). When no task is named, one Claude Opus 5.5 model adds modules for tasks it expects to be tested on over a small class-based language model, and the other is an n-gram model counted from 59,000 words it wrote. GPT-6.1 Sol's model uses exact rules for arithmetic and string tasks, and GPT-6 Astra's two models compile a hand-written knowledge base into sparse rule tables and hashed lookup tables over character pieces, with fixed arithmetic.

\begin{table}[t]
\centering
\small
\setlength{\tabcolsep}{4pt}
\resizebox{\linewidth}{!}{%
\begin{tabular}{lrrrrrr}
\toprule
LLM & Overall & ARC & MMLU & MathBench & SQuAD (F1) & CNN/DailyMail (ROUGE-L) \\
\midrule
\multicolumn{7}{l}{\textit{Four kinds of items named in the task}} \\
Claude Opus 5.5 & 33.6 & 45.0 & 30.0 & 30.2 & 34.0 & 24.6 \\
GPT-6 Astra & 28.1 & 32.0 & 26.5 & 27.2 & 30.2 & 23.7 \\
GPT-6.1 Sol & 27.7 & 30.4 & 28.9 & 24.9 & 29.0 & 23.6 \\
First three sentences (reference) & & & & & & 24.9 \\
\midrule
\multicolumn{7}{l}{\textit{No task named: write a language model}} \\
Claude Opus 5.5 & 26.1 & 37.2 & 26.8 & 25.8 & 15.2 & 13.7 \\
GPT-6 Astra & 19.1 & 25.1 & 24.0 & 25.5 & 3.0 & 0.8 \\
GPT-6.1 Sol & 18.6 & 24.2 & 23.0 & 24.4 & 3.9 & 1.6 \\
First three sentences (reference) & & & & & & 24.5 \\
\bottomrule
\end{tabular}}
\caption{\textbf{A single model written by Claude Opus 5.5 reaches 45.0\% on ARC, every other multiple-choice score stays within seven points of chance, and scores are lower when no task is named.} Each LLM row is the mean of two runs. \textit{Overall} averages ARC, MMLU, MathBench (accuracy in \%, chance 25), and the mean of SQuAD F1 and CNN/DailyMail ROUGE-L. The upper block evaluates 2,000 SQuAD items and 1,000 CNN/DailyMail articles, and the lower block evaluates 1,000 SQuAD items and 500 articles from the same sets.}
\label{tab:general_tasks}
\end{table}

\FloatBarrier

\section{Other analysis: text classification}\label{app:text}

On AG News topic classification (four classes, 7,600 test articles), Claude Opus 5.5 writes the weights of four standard architectures. These are a bag of words with 8,012 parameters, fastText with 32,100, a one-dimensional CNN with 33,732, and a small Transformer with 54,604. Each has two runs, and gradient descent uses Adam at learning rates 0.003, 0.001, and 0.0003 over three seeds, with a 2,000-label validation split for selecting the learning rate and step at every label count, including 100 labels. In the setting with no data, the writer sees only the four class names and the word list (the 2,000 most frequent training tokens) and sees no articles. Written in this setting, the four networks reach about 84\% test accuracy, which gradient descent from random initialization matches only with thousands of labeled articles (\cref{tab:agnews}); \textit{Labels GD needs to match it} is a log-linear interpolation over runs with 100, 1,000, 10,000, and 100,000 labels. With 100 labeled articles, starting from the written network keeps that accuracy while training from random initialization stays far below, showing that written weights can initialize gradient descent when labeled examples are scarce.

\begin{table}[t]
\centering
\small
\setlength{\tabcolsep}{4pt}
\resizebox{\linewidth}{!}{%
\begin{tabular}{lrrrrrr}
\toprule
 & Written, & Labels GD needs & \multicolumn{2}{c}{100 labels} & \multicolumn{2}{c}{10{,}000 labels} \\
\cmidrule(lr){4-5}\cmidrule(lr){6-7}
Architecture & no data & to match it & GD & Written, then GD & GD & Written, then GD \\
\midrule
Bag of words & 84.5 & 4{,}223 & 63.6 & 84.7 & 86.6 & 86.8 \\
One-dimensional CNN & 83.9 & 9{,}201 & 45.4 & 84.4 & 84.3 & 86.2 \\
fastText & 83.2 & 2{,}470 & 63.1 & 84.1 & 86.5 & 87.1 \\
Transformer & 83.7 & 8{,}191 & 49.5 & 84.4 & 84.5 & 86.0 \\
\bottomrule
\end{tabular}}
\caption{\textbf{A network written with no articles is worth thousands of labeled articles and is the better starting point when labels are few.} Accuracy columns report AG News test accuracy (\%), averaging two runs by Claude Opus 5.5 and three gradient-descent (GD) seeds started from random initialization or from the written network.}
\label{tab:agnews}
\end{table}

\FloatBarrier

\section{Other analysis: images, where WSP does not work}\label{app:images}

On CIFAR-10, chance is 10\%. Raw-pixel networks written by Claude Opus 5.5 reach 21--28\% test accuracy in all eight fixed architectures. These are an MLP of 201,482 parameters, a CNN of 24,234, a ViT of 75,082, and five easy-to-write designs of 1,666 to 10,658 parameters with named color and edge filters, region pooling, and score tables. This holds with no labels and with 192 images per class, where the writer sees three grids of 64 training images per class and may not process them with code or compute per-class statistics (\cref{tab:images}). A second frontier LLM, Claude Fable 5.2, reaches 28.8\% and 29.7\% on the CNN against 25.9\% and 24.1\% for Claude Opus 5.5. Which architecture is written best relates only weakly to which wins under gradient descent (Kendall's tau 0.21 to 0.29 at 45,000 labels).

Words and concepts have meanings that the LLM can state, whereas filter values at pixel positions do not. Grouping whitened $6\times 6$ pixel patches of 5,000 unlabeled training images into 128 k-means words, counted per image quadrant without a pretrained encoder, reaches 19.4\% with no labels and 29.7\% with 192 images per class when written, against 31.7\% with 100 labels and 58.7\% with 45,000 labels under gradient descent (\cref{tab:images}). When a self-supervised encoder turns each patch into the nearest of 256 centroids (spherical k-means on DINOv2-base patch features of 5,000 unlabeled training images), Claude Opus 5.5 writes a 2,570-parameter token-by-class score table. With no labels, the writer sees five training images containing each token, names the object in each picture, and tallies classes per token, reaching 96.1\%. Gradient descent on those tokens reaches 95.8\% with 100 labels, so the pretrained encoder accounts for the accuracy. A reconstruction tokenizer (VQGAN codes) does not help, since gradient descent with 10,000 labels reaches only 22.7--28.2\%. WSP as studied here therefore does not reach images and would need input units that an LLM can name without a pretrained encoder.

\begin{table}[t]
\centering
\small
\setlength{\tabcolsep}{4pt}
\begin{tabular}{lrrrr}
\toprule
 & \multicolumn{2}{c}{Written} & \multicolumn{2}{c}{Gradient descent} \\
\cmidrule(lr){2-3}\cmidrule(lr){4-5}
Input & No labels & 192 per class & 100 labels & 45{,}000 labels \\
\midrule
Raw pixels, eight architectures & 21.6--27.0 & 21.2--27.8 & 21.5--32.9 & 48.6--71.0 \\
6$\times$6 pixel-patch words & 19.4 & 29.7 & 31.7 & 58.7 \\
DINOv2 tokens & 96.1 & 96.2 & 95.8 & 98.0 \\
\bottomrule
\end{tabular}
\caption{\textbf{On raw pixels every written model stays far below gradient descent with all labels, while on DINOv2 tokens a written table matches gradient descent with 100 labels.} Cells give CIFAR-10 test accuracy (\%, chance 10) for Claude Opus 5.5 and gradient descent. Rows show ranges across eight architectures, single runs on patch words, and means on DINOv2 tokens. The no-label setting gives no ground-truth labels, and on DINOv2 tokens the writer looks at five unlabeled training images per token and names the objects itself.}
\label{tab:images}
\end{table}

\FloatBarrier

\end{document}